\documentclass[preprint,12pt]{elsarticle}

\usepackage{amssymb}
\usepackage{amsmath}
\usepackage{color}
\usepackage{physics}
\usepackage{algorithm,algpseudocode}
\usepackage{multirow}
\newcommand{\argmin}{\operatorname*{argmin}\limits}

\begin{document}

\begin{frontmatter}



\title{Physics-Informed Uncertainty Enables Reliable AI-driven Design}


\author[inst1]{Tingkai Xue}
\author[inst2,inst3]{Chinchun Ooi}
\author[inst4]{Yang Jiang}
\author[inst5]{Luu Trung Pham Duong}


\author[inst2]{Pao-Hsiung Chiu}
\author[inst2]{Weijiang Zhao}
\author[inst5]{Nagarajan Raghavan}
\author[inst6]{My Ha Dao}

\affiliation[inst1]{organization={Department of Mechanical Engineering, National University of Singapore},
            addressline={21 Lower Kent Ridge Road}, 
            city={Singapore},
            postcode={119007}, 
            state={Singapore},
            country={Singapore}}

\affiliation[inst2]{organization={Institute of High Performance Computing, Agency for Science Technology and Research},
            addressline={1 Fusionopolis Way}, 
            city={Singapore},
            postcode={138632}, 
            state={Singapore},
            country={Singapore}}

\affiliation[inst3]{organization={Centre for Frontier AI Research, Agency for Science Technology and Research},
            addressline={1 Fusionopolis Way}, 
            city={Singapore},
            postcode={138632}, 
            state={Singapore},
            country={Singapore}}

\affiliation[inst4]{organization={College of Electronics and Information Engineering, Shenzhen University},
            addressline={1211 Zhizhen Building}, 
            city={Shenzhen},
            postcode={518060}, 
            state={Guangdong},
            country={China}}

\affiliation[inst5]{organization={Engineering Product Design Pillar, Singapore University of Technology and Design},
            addressline={8 Somapah Road}, 
            city={Singapore},
            postcode={487372}, 
            state={Singapore},
            country={Singapore}}

\affiliation[inst6]{organization={Technology Centre for Offshore and Marine, Singapore},
            addressline={12 Prince George’s Park}, 
            city={Singapore},
            postcode={118411}, 
            state={Singapore},
            country={Singapore}}

\begin{abstract}
Inverse design is a central goal in much of science and engineering, including frequency-selective surfaces (FSS) that are critical to microelectronics for telecommunications and optical metamaterials. Traditional surrogate-assisted optimization methods using deep learning can accelerate the design process but do not usually incorporate uncertainty quantification, leading to poorer optimization performance due to erroneous predictions in data-sparse regions. Here, we introduce and validate a fundamentally different paradigm of Physics-Informed Uncertainty, where the degree to which a model's prediction violates fundamental physical laws serves as a computationally-cheap and effective proxy for predictive uncertainty. By integrating physics-informed uncertainty into a multi-fidelity uncertainty-aware optimization workflow to design complex frequency-selective surfaces within the 20 - 30 GHz range, we increase the success rate of finding performant solutions from less than 10\% to over 50\%, while simultaneously reducing computational cost by an order of magnitude compared to the sole use of a high-fidelity solver. These results highlight the necessity of incorporating uncertainty quantification in machine-learning-driven inverse design for high-dimensional problems, and establish physics-informed uncertainty as a viable alternative to quantifying uncertainty in surrogate models for physical systems, thereby setting the stage for autonomous scientific discovery systems that can efficiently and robustly explore and evaluate candidate designs.

\end{abstract}




\end{frontmatter}


\section{Introduction}
\label{sec:intro}
Inverse design is a ubiquitous problem statement spanning multiple disciplines in science and engineering from drug discovery and materials science to electrical engineering \cite{xie2023inverse, li2022empowering, yang2023inverse}. Historically, progress has been driven by a combination of expert intuition and iterative, costly physical prototyping, a process that inherently limits the scope of exploration and hinders the discovery of novel, non-intuitive solutions. The rise of computational science offered a paradigm shift, enabling virtual testing that is more resource-efficient than physical fabrication. However, the computational cost of these models still render exhaustive search in vast, high-dimensional design spaces intractable. To overcome this bottleneck, data-driven surrogate models, particularly deep neural networks, have emerged as a powerful solution, capable of predicting performance in milliseconds and dramatically accelerating the design optimization process \cite{ma2021deep, naseri2021combined, jiang2020surrogate}. 


Crucially, these data-driven models are inherently unreliable when extrapolating. Lacking any grounding in physical principles, they can produce confidently incorrect predictions in data-sparse regions of the design space, which engenders significant brittleness when incorporated in optimization algorithms for inverse design \cite{chatterjee2019critical, li2019surrogate}. In being guided by these surrogates, one can be deceptively led into false minima during the automated design optimization process, i.e. regions that appear optimal to the model but are actually poor-performing. The conventional approach to mitigate this risk is through statistical uncertainty quantification. Bayesian methods and certain forms of statistical learning algorithms (e.g. Gaussian processes) are prized for their ability to provide estimates of uncertainty in their predicted output. Similarly, Bayesian neural networks have been proposed as a way to incorporate probabilistic estimation or uncertainty in deep learning models, while Deep Ensembles are another simple yet effective alternative \cite{lakshminarayanan2017simple,tran2019bayesian,wenzel2020hyperparameter}. Specific to design optimization, uncertainty estimates are a key component of any strategy to balance the classic `exploration-exploitation' trade-off. Uncertainty estimates can guide the algorithm to `explore' novel regions where the model is least confident, while the model's predictions guide it to `exploit' known regions of high performance \cite{xiong2019data, jung2021confidence}. While effective, these methods can be computationally burdensome and, more fundamentally, do not leverage the rich, underlying physical laws that govern the system. 

\begin{figure}[!t]
	\begin{center}
		{\includegraphics[width=0.9\columnwidth]{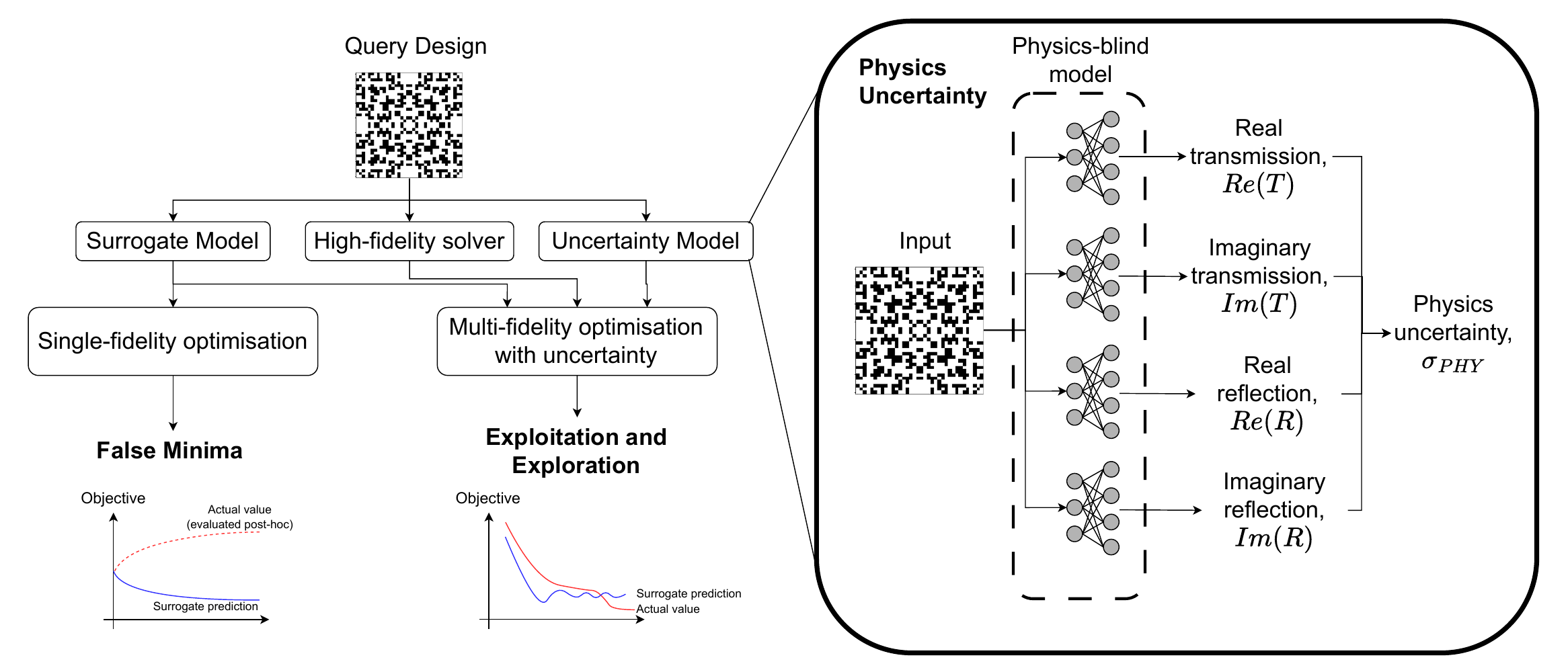}}
            \caption{Illustration of how a single-fidelity surrogate-assisted optimization workflow may lead to erroneous designs as the design deviates from its' training distribution whereas an uncertainty-aware multi-fidelity workflow efficiently utilizes high-fidelity evaluations in balancing `exploration' and `exploitation' for robust design optimization. The novel aspect of this work is the validation of physics-based rules as an effective uncertainty metric in this real-world problem of metasurface design.}
		\label{Fig:graphical_abstract}
	\end{center}
\end{figure}

Here, we introduce and validate a fundamentally different paradigm: Physics-Informed Uncertainty. Simply, a surrogate model’s credibility can be directly, cheaply, and effectively assessed by quantifying the degree to which its predictions violate fundamental physical laws. By evaluating a model’s compliance with principles like continuity in electromagnetic field or conservation of energy, we can generate a powerful, low-cost proxy for predictive uncertainty. This approach is not only computationally efficient but also versatile, allowing it to be integrated with any pre-trained surrogate model to guide optimization. To rigorously test this principle, we apply it to the inverse design of frequency-selective surfaces (FSS), an especially challenging problem due to its extremely high-dimensional design space ($\approx 10^{97}$ possibilities), and great practical importance in telecommunications and optics as a fundamental building block of metasurfaces \cite{li2018metasurfaces}. An illustrative schematic of this physics-informed uncertainty metric and its potential utility in robust uncertainty-aware multi-fidelity design optimization is in Fig \ref{Fig:graphical_abstract}.

We demonstrate that a standard surrogate-assisted particle swarm optimization workflow catastrophically fails, achieving a success rate of less than 10\%. In contrast, a multi-fidelity uncertainty-aware framework, guided by the physics-informed uncertainty, is able to more robustly find performant solutions at $\approx 5x$ the rate, even while maintaining a 10-fold reduction in computational cost relative to the sole use of high-fidelity numerical solvers. While there is some prior literature on the use of machine learning for the design of frequency selective surfaces, we note that this is the first work, to our knowledge, to both demonstrate the potential drawbacks, especially in such high-dimensional spaces whereby no practical dataset is anticipated to ever be fully comprehensive, and to incorporate ideas of uncertainty into the design optimization workflow\cite{calik2021accurate,fontoura2021synthesis,genovesi2006particle}. More broadly, the ability to estimate uncertainty using physical principles introduces a promising direction for domain experts to encode prior knowledge as constraints or indicators of model reliability, establishing a generalizable strategy for reliable and efficient AI-driven scientific discovery.

\section{Methods}
\label{sec:method}


\subsection{Description of frequency selective surface}

Frequency-selective surfaces are a fundamental component of many complex electronic devices, including in advanced telecommunications \cite{katwe2024overview, chen2024metasurfaces}. For this work, we focus on the performance of devices in the 20 - 30 GHz range, which is crucial to 5G and sub-6G communication systems. Crucially, this is currently the most challenging frequency range that is manufacturable. Current fabrication technologies limit the design space to simple shapes for frequencies much higher than this range, while the dimensionality of the problem for frequencies in the MHz and below range is much lower and can be well-studied by conventional approaches. 

For this frequency range, a typical meta-atom (basic building block in the fabrication of a frequency-selective surface) is defined on a $5.4 \times 5.4 mm^2$ planar area. While the meta-atom design can theoretically be any free-form composition of free-space and a metallic conductor, it is common to utilize a pixel-based discretization to parameterize the entire design space, which determines the dimensionality of the design problem. In pursuit of high-performance designs, cells with more complex patterns are increasingly desired, but this inadvertently increases the design space. The fundamental choice of number of pixels needs to balance the ability to produce sufficiently complex designs while satisfying practical constraints like maintaining a reasonably-sized search space and ease of fabrication during scale-up.  

\begin{figure}[!t]
	\begin{center}
		{\label{group1}\includegraphics[width=0.9\columnwidth]{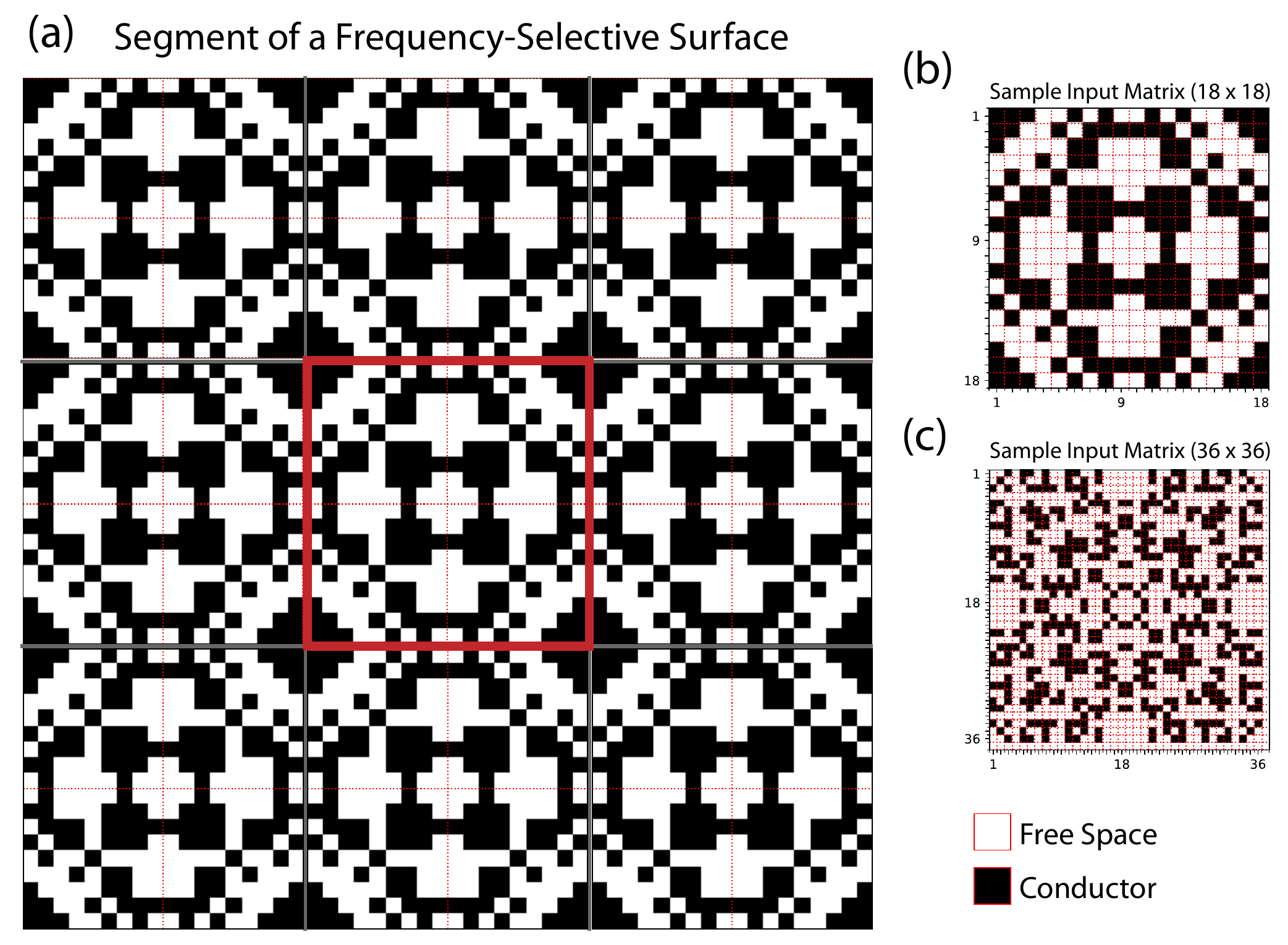}}
		\caption{(a) Illustration of how a periodic arrangement of a meta-atom is frequently used to construct a frequency-selective surface and other useful devices. Segment in the red box is the periodic meta-atom. (b) and (c) represent different possible designs that can be constructed via an 18×18 or 36×36 pixel discretization of the design space of a meta-atom.}\label{Fig:Metasurface}
	\end{center}
\end{figure}

Based on prior experience, an $18 \times 18$ pixel discretization is chosen as this best balances ease of fabrication and the complexity of the high-dimensional design space for this physical system. In addition, it is worth noting that a meta-atom with an $18 \times 18$ pixel parameterization, as illustrated in Figure~\ref{Fig:Metasurface}, still has $\approx 10^{97}$ different possible designs. Hence, additional constraints are commonly placed on the design, such as eight-fold symmetry. The symmetrical design ensures good angular stability of the metasurface independent of incident angle and polarization while reducing the design space substantially. In this instance, the design space is now reduced to $2^{45}$ designs (which is still $\approx$ 35 trillion possible designs). As the design space is extremely high-dimensional, design of experiment (DOE) approaches or other optimization algorithms need to iterate through multiple potential designs. 

It is worth noting that the high-dimensional design space characteristic of this domain naturally precludes any accumulated dataset from ever being fully representative of the entire design space, further motivating the use of uncertainty-aware measures during the design process. Indeed, this relative prediction uncertainty is hypothesized to be critical for many real-world problems beyond frequency-selective surfaces \cite{miller2019predictive, zhao2022limitations}, especially for design problems where new and potentially novel designs are sought and comprehensive datasets may inherently not contain the novel designs desired.


\subsection{Data generation}
The performance of metasurface designs are typically evaluated with a high-fidelity numerical simulator. These numerical solvers solve the fundamental Maxwell's Equations in order to yield information about their transmission and reflection responses in the wavelength range of interest. However, conventional high-fidelity numerical solvers are extremely computationally expensive, motivating the development and use of faster models. 


Commercial electromagnetic solvers (e.g. HFSS) are frequently employed to evaluate the frequency responses of meta-atom designs. An automation script was developed to utilize an electromagnetic solver to a) create a physical model,  b) assign master-slave boundary conditions, c) define Floquet ports, and d) solve the transmission and reflection coefficients under co- and cross-polarization in the 20 - 30 GHz range to constitute a starting high-fidelity dataset. The high-fidelity numerical solver takes between one and ten minutes to solve each meta-atom design to a convergence criteria of 0.01. In total, 10,000 meta-atom designs with eight-fold symmetry are generated randomly and evaluated over a period of approximately two months. 

\begin{figure}[!t]
	\begin{center}
		{\label{group1}\includegraphics[width=0.9\columnwidth]{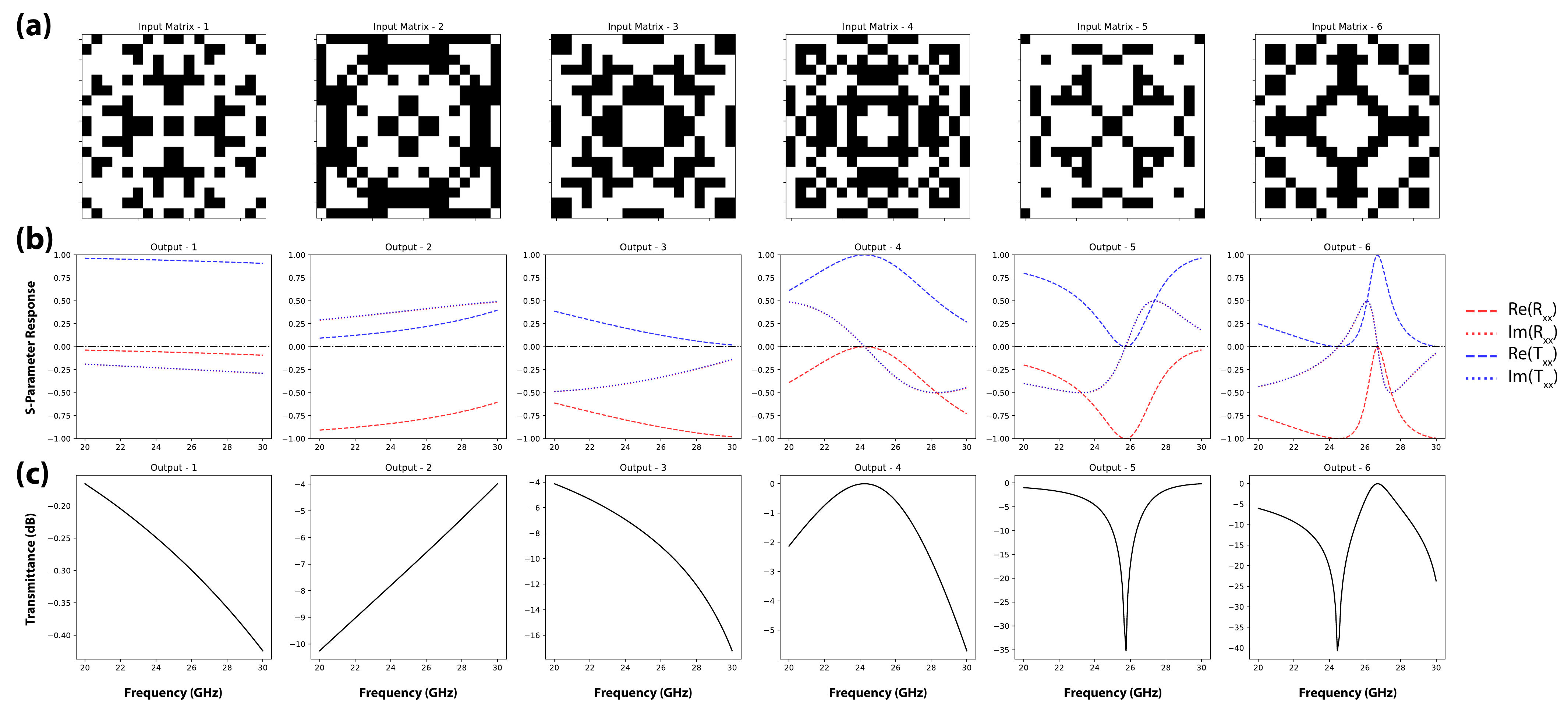}}
		\caption{(a) Illustration of 6 examples of (a) input design and their corresponding (b) S-parameter responses and the (c) Transmission (in dB) within the 20 - 30 GHz spectrum.}\label{Fig:HFSS_example}
	\end{center}
\end{figure}

Sample plots of the 4 response curves obtained by the high-fidelity solver for various input designs are in Figure~\ref{Fig:HFSS_example}. These curves are the S-parameter responses corresponding to the Real (Re) and Imaginary (Im) part of the Transmission (T) and Reflection (R) spectrum for \textit{xx-}polarization. As described previously, a common simplifying constraint is that the design should be symmetrical, thereby ensuring little cross-polarization. Under this constraint, the \textit{xx}- and \textit{yy}- polarizations are essentially identical.

In addition, Figure~\ref{Fig:HFSS_example} shows the variety of responses that can be produced by various designs, spanning responses that exhibit monotonic behavior and resonant responses with bandstop-type (e.g. 5th design) or bandpass-type behavior in the frequency spectrum of interest. In very rare instances, a double resonance peak can also be observed in the dataset, as illustrated in the 6th example.

\subsection{Neural network to predict frequency selective surface response}

The generated dataset with $10,000$ designs is subsequently used to train a deep learning model that can be incorporated into a design optimization workflow as a replacement for the high-fidelity numerical model. As per the numerical model, the input to the neural network model remains the $18 \times 18$ bit matrices of generated meta-atom patterns, while the outputs are the real and imaginary parts of transmission and reflection across the surface. 

\begin{figure}[!t]
	\begin{center}
		{\label{group1}\includegraphics[width=0.9\columnwidth]{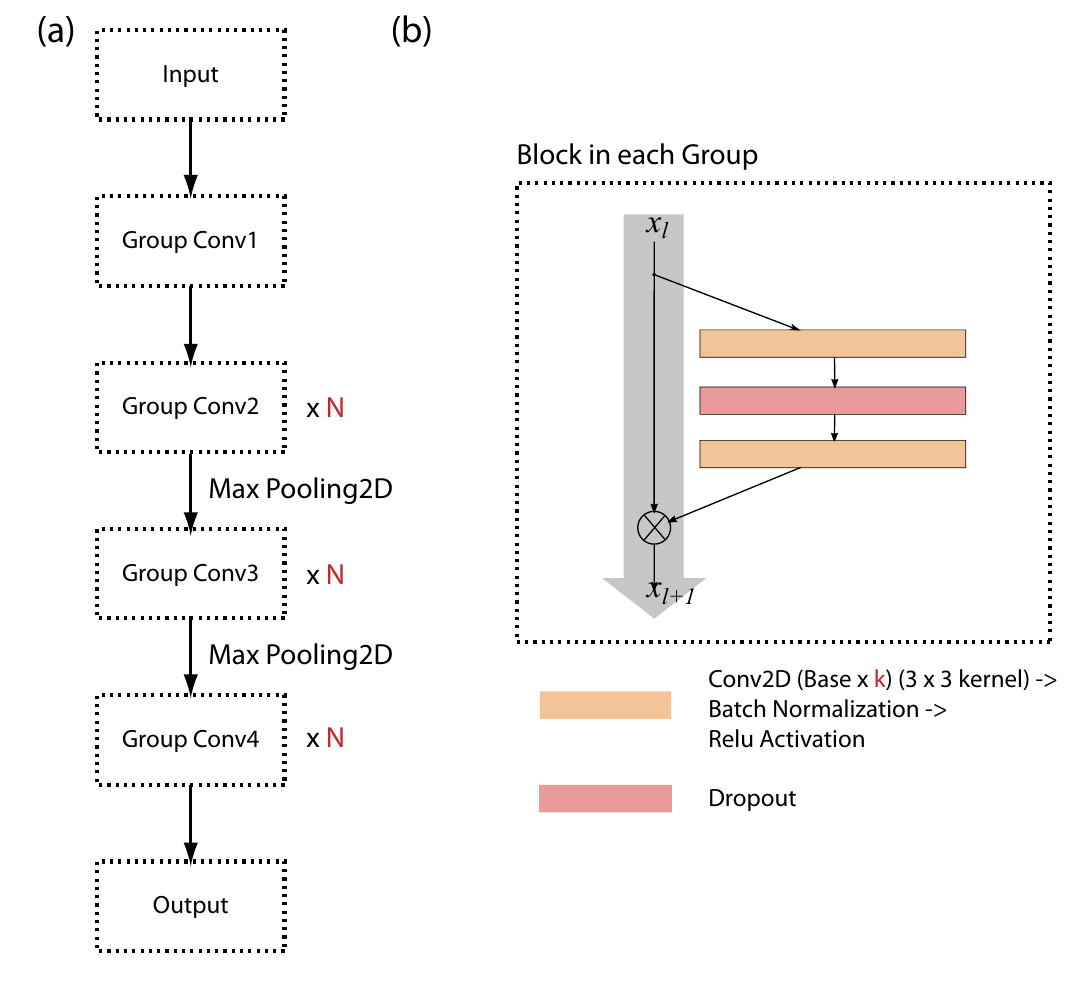}}
		\caption{(a) Schematic of the WideResNet architecture that was used for the prediction of S-parameter responses based on the input matrices. (b) Illustration of the operations within each group illustrated in (a). }\label{Fig:WideResNet}
	\end{center}
\end{figure}

Briefly, the full dataset was randomized into training, validation and test datasets in an $8:1:1$ fraction. The training and validation sets were passed through a neural architecture search pipeline to derive the best neural network architecture and hyper-parameters. A WideResNet architecture was finally chosen as it consistently outperformed the other choices tested during this process. Network hyper-parameters such as the depth, and number of layers were also selected based on their performance on the validation set. A schematic of the chosen architecture is presented in Figure~\ref{Fig:WideResNet}. In keeping with the basic architecture and nomenclature outlined in the original WideResNet work~\cite{zagoruyko2016wide}, the final architecture used also comprises 3 blocks (Group Conv2, Conv3 and Conv4) in Figure~\ref{Fig:WideResNet}(a) which are repeated $N$ times. The groups comprise of 2 Conv2D layers, followed by a Batch Normalization and a Relu activation layer, separated by a Dropout layer. The Base variable in \ref{Fig:WideResNet}(b) are 16, 32, and 64 respectively for Conv2, Conv3 and Conv4. The $k$ variable is a multiplier that is tuned along with $N$ during hyper-parameter optimization. Additional description of the chosen deep learning surrogate model is in ~\cite{yang2022physics}. 

\subsection{Uncertainty estimates of neural network prediction}
Deep learning models such as the WideResNet typically only provide predictions for the performance metric of interest (i.e. the transmission and reflection profile in this work). For this work, we further estimated the relative predictive uncertainty via two methods, i) an ensemble-based estimation, and ii) a physics-based estimation~\cite{yang2022physics}. 

Deep ensembles have previously been shown to be a simple yet effective means of estimating predictive uncertainty from deep learning models \cite{lakshminarayanan2017simple,tran2019bayesian,wenzel2020hyperparameter}. The basic intuition is that different deep learning models, when trained with the same dataset, should converge to a similar prediction in data-rich portions of the training data distribution. Hence, large discrepancies across predictions from different models correspond to a greater likelihood that the individual model predictions are wrong. This approach is the first method employed (ENSB-UNC). 

\begin{equation}
        \text{ENSB-UNC}=\sigma_{ENSB}=\sqrt{\frac{\sum_{i}^{N_m}\left\|f_i-\bar{f}\right\|^2}{N_m}}, \text{ where } \bar{f}=\frac{1}{N_m}\sum_{i}^{N_m} f_i
\end{equation}

For this work, $N_m$ = 10 models were used, and the function, $f_i$, represents the transmission profile predicted by each of the models. The transmission profile is predicted at 100 uniformly spaced points spanning 20 - 30 GHz.

In previous work, physics uncertainty was proposed as an effective metric to quantify confidence in a deep learning model's prediction~\cite{yang2022physics}. Intuitively, the larger the extent to which physics rules are being violated by a model's prediction, the more likely it is that the model prediction is wrong. For this particular design problem, the following equations need to be obeyed for both reflection, $R$, and transmission, $T$, as a consequence of the necessary boundary condition for electromagnetic fields which dictates that the tangential component of the electromagnetic field needs to be continuous across the metasurface:
\begin{subequations}
    \begin{align}
        1+Re(R)&=Re(T)\\
        Im(R)&=Im(T)
    \end{align}
\end{subequations}

The magnitude of physics uncertainty (PHY-UNC) can be defined as the sum of the degree of violation (residuals) of both equations. This also bears similarity to how physics-informed losses are computed in typical physics-informed neural networks \cite{karniadakis2021physics, raabe2023accelerating}, although physics-informed losses are inserted to improve physics compliance during training, whereas this work proposes to utilize this as a \textit{post-hoc} criteria for assessing the performance of a purely data-driven model.
\begin{subequations}
    \begin{align}
        \text{PHY-UNC}=\sigma_{PHY}(x)&=l_{Re}(x)+l_{Im}(x)\\
        l_{Re}(x)&=|Re(T(x))-Re(R(x))-1|\\
        l_{Im}(x)&=|Im(T(x))-Im(R(x))|
    \end{align}
\end{subequations}

Critically, only one version of the model is required, in contrast to the multiple models ($N_m$) required for evaluating ENSB-UNC. However, this model needs to be trained to predict both the real and imaginary transmission and reflection in order to evaluate PHY-UNC.

\subsection{Design optimization workflow for frequency selective surface}

Particle swarm optimization (PSO) algorithms are frequently used for design optimization across many engineering problems, including microelectronics \cite{wang2024designing,lalbakhsh2016multiobjective}. They can be easily integrated into any black-box surrogate-based optimization framework with no requirements on differentiability, but face significant limitations with regards to computational cost as they require repeated calls to a black-box simulator (e.g. high-fidelity numerical solver or low-fidelity deep learning surrogate model) during their search. 

The binary version of PSO (BPSO) was previously proposed by Kennedy and Eberhart, and is particularly suitable for the current problem where the design variables are a $18 \times 18$ binary matrix~\cite{kennedy1997discrete}. The main idea of BPSO is to use a transfer function to map the continuous search space into a binary one, e.g. via an S-shaped transfer function~\cite{mirjalili2013s}. Critically, the inherent versatility of the BPSO algorithm allows easy integration with both a high-fidelity solver and the developed surrogate (with and without the additional uncertainty estimation) for our selected design tasks. For ease of integration, we use an in-house PEEC solver which has been previously validated and shown to be as accurate as commercial solvers~\cite{jiang2022full, jiang2022compact, jiang2023peec}

\subsubsection{Design towards band-stop and band-pass surfaces}
\begin{figure}[!t]
	\begin{center}
		\includegraphics[width=1.0\columnwidth]{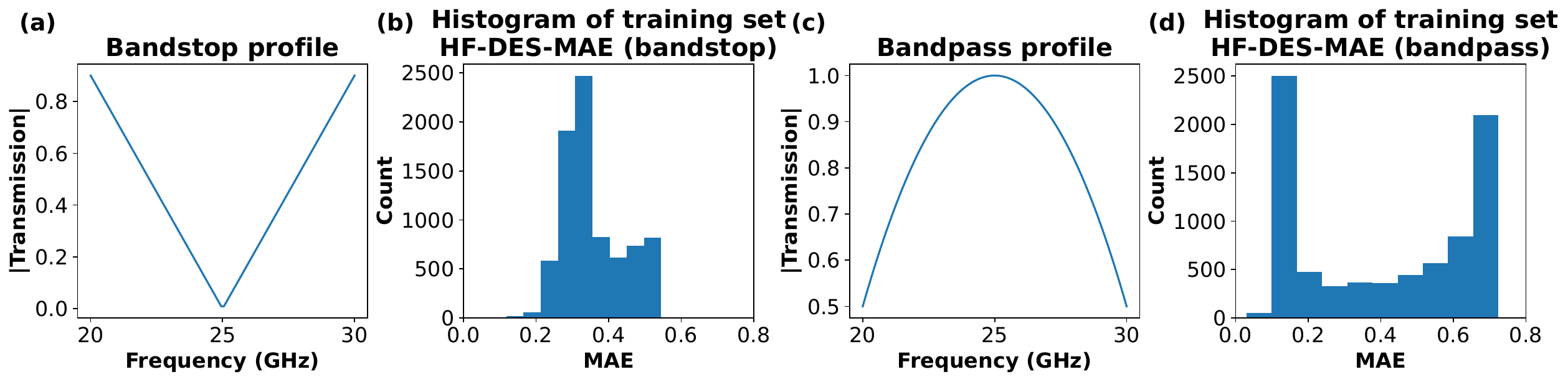}
		\caption{Band-pass and band-stop profiles and the distributions of training set HF-DES-MAE with respect to the profiles \label{fig:profile_hist}}
	\end{center}
\end{figure}

As a demonstration, we target the design of both a band-pass and band-stop surface in the 20 - 30 GHz range. The target transmission profiles used are shown in Figure~\ref{fig:profile_hist}(a) and (c). 

In this context, it is useful to explicitly define the following metrics: 
\begin{subequations}
    \begin{align}
        \text{LF-MAE} =\varepsilon_{LF}(x)&= |f_{LF}(x)-f_{HF}(x)|\\
        \text{LF-DES-MAE} =\Delta_{LF}(x)&= |f_{LF}(x)-f_{design}|\\
        \text{HF-DES-MAE}=\Delta_{HF}(x) &= |f_{HF}(x)-f_{design}|
    \end{align}
\end{subequations}

Hence, LF-MAE quantifies the accuracy of the surrogate model itself via the mean absolute error (MAE) between the low-fidelity surrogate model predictions $(f_{LF})$ and the corresponding high-fidelity ground truth result from a numerical solver $(f_{HF})$. In contrast, LF-DES-MAE and HF-DES-MAE quantifies the distance between the predicted and target transmission profiles via the difference (MAE) between the desired performance profile $(f_{design})$ and the low-fidelity surrogate model prediction and the high-fidelity solver result (ground truth) respectively.

Distributions of the 10,000 samples' HF-DES-MAE with respect to both transmission profiles are plotted in Figure \ref{fig:profile_hist}(b) and (d). This distribution is a proxy for the difficulty of each design task in this design space. We first note that these two design problems can be approximately satisfied by multiple possible designs (an ill-posed inverse design problem). In enumerating the discrepancy between randomly generated samples and the desired target spectrum, a larger fraction of random designs producing responses similar to the desired target spectrum is indicative of the design space containing more possible solutions. This higher prevalence of random designs with a specific desired response also suggests that the design task is `simpler', as the BPSO is more likely to be initialized near a potential, satisfactory solution and should converge more easily.

For these two targets, it is clear that more of the 10,000 samples previously randomly generated have lower HF-DES-MAE (around $0.1$) for the band-pass profile than the band-stop profile, for which a majority of the training examples have a HF-DES-MAE around $0.3$. This suggests that the band-pass profile is more common in the design space (and hence, the training dataset), with significant implications for the surrogate-assisted design optimization process (further discussed in Section \ref{sec:uncertainty-awareBPSO_dis}). 

\subsubsection{Design with baseline BPSO}
As a baseline, a conventional implementation of the BPSO algorithm using an S-shape transfer function was used. In each iteration, each particle's velocity is updated towards its own particle best and a global best. In BPSO, the velocities relate to the probabilities of generating 1s and 0s at each coordinate, from which the particles' positions are updated. The particle bests are updated based on the LF-DES-MAE predicted by the surrogate model while the global best is the particle with the lowest LF-DES-MAE thus far. When only the surrogate model is used to select the global and particle bests, this implementation of BPSO is a single-fidelity approach. The detailed algorithm is described in Algorithm \ref{algo:vanilla_bpso} while an illustrative flow-chart is presented in \ref{sec:flowcharts}.

\begin{algorithm} 
\caption{Baseline Binary Particle Swarm Optimization}\label{algo:vanilla_bpso}
\begin{algorithmic}[1]
\Require $N_{itr}$: number of iterations; $N$: number of particles; $d$: dimension of each particle (equals to 45 for one octant of a $18\times 18$ design); $\Delta_{LF}$: function that returns LF-DES-MAE; $c_{1},c_{2}$: acceleration coefficients due to particle and global bests respectively; $w$: inertial coefficients

    \State $P_{gbest}^{itr} \gets \text{empty list}$
    \State Initialize particles $X=[x^1,...,x^N]\in \{0,1\}^{N\times d}$
    \State Initialize velocities $V\in \mathcal{R}^{N\times d}$
    \State $p_{best} \gets X$
    \State $g_{best}\gets \underset{p\in p_{best}}{\arg \min} \Delta_{LF}(p)$
    \For{$itr\in \{1,...,N_{itr}\}$}
        \For{all particles $x^i\in X$}
            \State $ V^i\gets wV^i+c_{1}r_1(p_{best}^i-x^i)+c_{2}r_2(g_{best}-x^i)$
            \State $S^i\gets \frac{1}{1+e^{-3V^i}}$
            \State $x^i\gets (S^i>rand(d))$
            \State $p_{best}^i\gets \argmin_{p\in \{p_{best}^i, x^i\}} \Delta_{LF}(p)$
        \EndFor
    
        \State $g_{best}\gets \underset{p\in p_{best}}{\arg \min} \:\Delta_{LF}(p)$
        \State Add $g_{best}$ to $P_{gbest}^{itr}$
    \EndFor
\State \textbf{return} $P_{gbest}^{itr}$
\end{algorithmic}
\end{algorithm}

\subsubsection{Design with uncertainty-aware BPSO} \label{sec:uncertainty-awareBPSO_dis}
\begin{figure}
    \centering
    \includegraphics[width=1.0\linewidth]{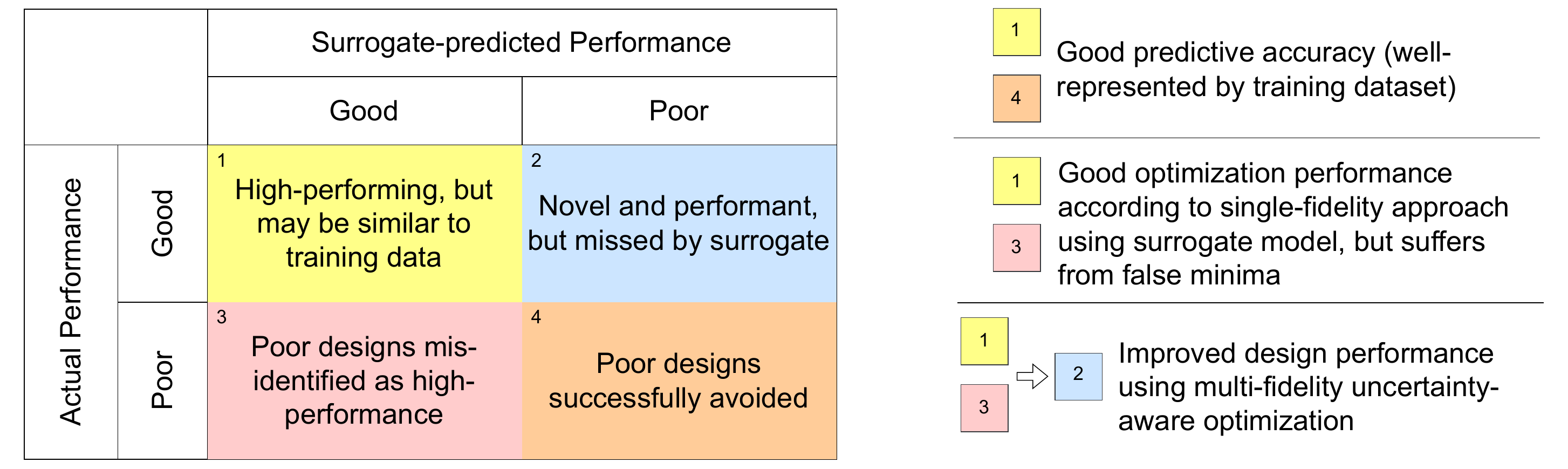}
    \caption{Illustrative example of different regions of the design space}
    \label{fig:design_space_quadrant}
\end{figure}

In high-dimensional design spaces, the selected design optimization algorithm may converge to a non-performant design (region colored red in Figure \ref{fig:design_space_quadrant}) if the guidance by the surrogate model is inaccurate, an especially important issue for novel designs. In addition, exploring outside the training set (region colored blue in Figure \ref{fig:design_space_quadrant}) is particularly necessary if the training set inherently contains very few designs close to the desired target, such as for the band-stop profile. 

To demonstrate these problems, a modification is proposed to the baseline BPSO algorithm. In the original BPSO, if the global best selected is a particle with a putatively low but inaccurate LF-DES-MAE, the global best will be retained throughout the run and particles may end up gathering around this false minima. Therefore, instead of biasing each particle's search with a single global best, each particle now only tracks its own best LF-DES-MAE during each iteration. Hence, this ablation study allows to empirically show that the optimization process can be inadvertently derailed by such `falsely optimal designs'. The modified algorithm is detailed in Algorithm \ref{algo:single_pbest}. 

More generally, multiple options can be used for evaluation of the particle best. When all particles are evaluated using the high-fidelity solver during the optimization (HF-DES-MAE), this gives the most accurate estimation of the particle best, and is expected to yield the best possible optimization performance (with the trade-off being the accompanying higher computational cost). On the contrary, the sole use of a low-fidelity surrogate model (LF-DES-MAE) will yield challenges similar to the baseline BPSO. 

The choice of particles with high uncertainty is also an option that may be helpful. When the performant design is rare or absent in the training set distribution, biasing the particle velocity towards regions with higher uncertainty will enhance exploration and naturally produce designs outside of the training set distribution (which may now be performant). Hence, the selection of particle bests by assessing the relative predictive uncertainty to increase exploration is also tested. This metric can be either the physics-based uncertainty (PHY-UNC) or the ensemble-based uncertainty (ENSB-UNC).  

\begin{algorithm} 
\caption{Attraction to metric-defined particle best}\label{algo:single_pbest}
\begin{algorithmic}[1]
\Require $N_{itr}$: number of iterations; $N$: number of particles; $d$: dimension of each particle (equals to 45 for one octant of a $18\times 18$ design); $\Delta_{LF}$: functions that return LF-DES-MAE; $m\in\{\sigma_{PHY}, \sigma_{ENSB},\Delta_{LF},\Delta_{HF}\}$: function that computes a specified metric (PHY-UNC, ENSB-UNC, LF-DES-MAE, HF-DES-MAE);$c_{metric}$: acceleration coefficients due to metric-defined particle best; $w$: inertial coefficients

    \State $P_{eval}^{itr} \gets \text{empty list}$
    \State Initialize particles $X=[x^1,...,x^N]\in \{0,1\}^{N\times d}$
    \State Initialize velocities $V\in \mathcal{R}^{N\times d}$
    \State $p_{best} \gets X$
    \State $p_{metric}\gets m(p_{best})$
    \For{$itr \in \{1,...,N_{itr}\}$}
          \For{all particles $i=1...N$}
                \State $ V^i\gets wV^i+c_{metric}r(p_{best}^i-x^i)$
                \State $S^i=\frac{1}{1+e^{-3V}}$
                \State $x^i\gets (S^i>rand(d))$
                \State $ p_{best}^i\gets \argmin_{p\in \{p_{best}^i, x^i\}} m(p)$
                \State $p_{metric}^i\gets m(p_{best}^i)$
            \EndFor

      \State $x^{eval}\gets \underset{x\in X}{\arg \min}\:\Delta_{LF}(x)$
      \State Add $x^{eval}$ to $P_{eval}^{itr}$
\EndFor

\State \textbf{return $P_{eval}^{itr}$}

\end{algorithmic}
\end{algorithm}

\subsubsection{Design Performance with a Multi-fidelity Uncertainty-aware BPSO} \label{sec:multifidelity_bpso}
In the previous Sections, particles are evaluated using a high-fidelity numerical solver post-hoc, after the optimization search had concluded. If this high-fidelity re-evaluation is performed during the iteration itself, this partial knowledge of the accurate objective function profile can be exploited to more robustly guide the particles to a better solution without drastically increasing the number of costly high-fidelity evaluations. This is the bedrock of many multi-fidelity optimization approaches, where high-fidelity evaluations are intertwined with low-fidelity evaluations to improve performance, even while keeping within a limited evaluation budget. 

Uncertainty estimates are key to reducing maximum MAE under a limited computational budget. In this process, one might use a fast, low-fidelity surrogate model to predict the performance across many possible designs, and selectively re-evaluate less confident predictions with a more computationally expensive (and slower) high-fidelity numerical solver. Experiments detailed in \cite{yang2022physics} showed that the selection of $10\%$ of the test set for high-fidelity re-evaluation via uncertainty metrics effectively reduced the maximum test MAE by $50-60 \%$. On the other hand, randomly picking designs for high-fidelity re-evaluation is highly ineffective and has negligible impact on reducing the maximum test MAE. The use of uncertainty metrics to search and select samples for re-evaluations under a constrained evaluation budget can greatly improve robustness against erroneous convergence to non-performant designs

The proposed multi-fidelity integration in this work is implemented as a combination of two distinct stages. The first, termed the `constant-attraction' stage, maintains a single optimization objective throughout the search (similar to Algorithm \ref{algo:single_pbest}), and shares commonalities to methods used in Bayesian optimization to overcome the `cold-start' problem. At each iteration, a particle is selected for re-evaluation by the high-fidelity numerical solver according to a pre-defined metric, and this set of evaluations are used in the next stage. During the `alternating-strategy' stage, detailed in Algorithm \ref{algo:alternation} and Figure \ref{Fig:algorithm_flowchart_3} in the Appendix, the optimization algorithm dynamically shifts between exploiting known performant regions and exploring regions of the design space with greater uncertainty, as per common implementations of the `exploit-explore' paradigm. 

Hence, the alternating approach involves the following velocity update. 
\begin{equation}
        V^i\gets wV^i+c_{explore}r_1(p_{best}^i-x^i)+c_{exploit}r_2 F_{att}(B, x^i)
\end{equation}
where $p_{best}^i$ is the particle best in terms of highest uncertainty and 

\begin{equation}
    \begin{aligned}
        F_{att}(B,x^i)&=\frac{1}{\sum_j \alpha_j}\sum_{j} \alpha_j (b^j-x^i)\\
        \text{ where } B&=\{b^j|1\leq j\leq N\}, \alpha_j=e^{-k\cdot \Delta_{HF}(b^j)}
    \end{aligned}
\end{equation}

Fundamentally, the `alternating-strategy' stage is characterized by exploration, which is achieved by guiding particles toward regions of high uncertainty, as quantified by $c_{explore}r_1(p_{best}^i-x^i)$, and exploitation, which is achieved by guiding particles towards regions of high performance, as quantified by $c_{exploit}r_2 F_{att}(B, x^i)$. During each iteration, all designs previously re-evaluated with the high-fidelity model are used as `beacons' for the particles, akin to attractors in dynamical systems \cite{hawkins2021attractors}. An attraction factor $\alpha_j$ is calculated based on its HF-DES-MAE and the velocity update of each particle $x^i$ will be a sum of its difference vector with every beacon in the set of beacons $B$, scaled by the attraction factor. This ensures that the algorithm is able to leverage valuable information from the costly and sparse high-fidelity evaluations during the optimization process.

\begin{figure}[H]
	\begin{center}
		{\label{group1}\includegraphics[scale=0.9]{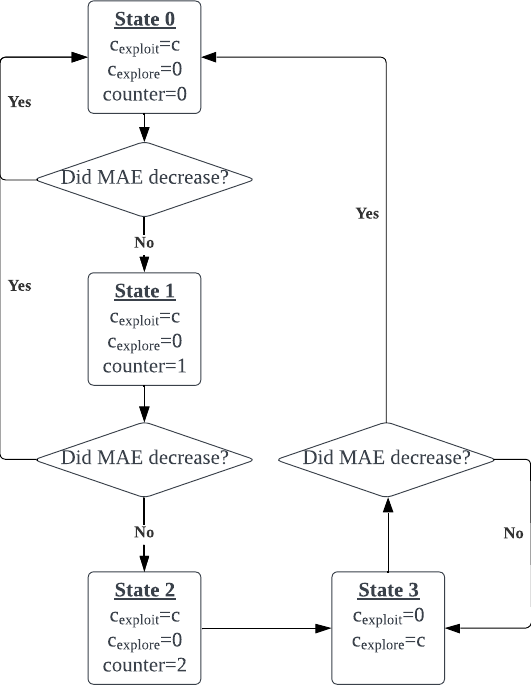}}
		\caption{State diagram of alternation between exploitation and exploration during the `alternating-strategy' stage of Algorithm \ref{algo:alternation}}\label{Fig:alternating_state}
	\end{center}
\end{figure}

Doing exploration and exploitation simultaneously may lead to conflicting updates and hinder convergence. Hence, particles alternate between exploration and exploitation. Initially, particles are attracted to the beacons. However, when the lowest HF-DES-MAE has not decreased for two iterations, they alternate to the particle best in terms of highest uncertainty until a decrease in HF-DES-MAE is observed. The state transition diagram is illustrated in Figure \ref{Fig:alternating_state}. For this work, the multi-fidelity optimization used 10 iterations of the `constant-attraction' stage before switching to 20 iterations of the `alternating-strategy' stage. 

\begin{algorithm} 
\caption{Multi-fidelity Uncertainty-aware BPSO}\label{algo:alternation}
\begin{algorithmic}[1]
\Require $N_{itr}^{const},N_{itr}^{alter}$: number of iterations of constant and alternating stages respectively;  $N$: number of particles; $d$: dimension of each particle (equals to 45 for one octant of a $18\times 18$ design); $f_{LF}, f_{HF}$: functions that return LF-DES-MAE and HF-DES-MAE respectively; $f_{metric}$: function that computes a specified metric - PHY-UNC or ENSB-UNC; $f_{att}$: function that computes the attraction vector from set of beacons to a particle ; $c$: acceleration coefficient; $w$: inertial coefficient

    \State Perform $N_{itr}^{const}$ iterations of constant stage
    \State $B^{itr} \gets \text{set of HF-evaluated particles during the constant stage}$
 \State $\varepsilon^{itr} \gets \text{HF-DES-MAE of particles in } B^{itr}$
\State $X \gets \text{set of particles after the constant stage}$
\State $V \gets \text{velocities after the constant stage}$
\State $c_{exploit} \gets c; c_{explore} \gets 0$
\State $counter \gets 0$
    \For {$itr \in \{1,...,N_{itr}^{alter}\}$}
        \For{all particles $i=1...N$}
            \State $ V^i\gets wV^i+c_{explore}r_1(p_{best}^i-x^i)+c_{exploit}r_2 f_{att}(B^{itr}, x^i)$
        \State $S^i=\frac{1}{1+e^{-3V^i}}$
        \State $x^i\gets (S^i>rand(d))$
        \State $ p_{best}^i\gets \underset{p\in \{p_{best}^i, x^i\}}{\arg\min} f_{metric}(p)$
        \EndFor

        \State $x^{eval}\gets \underset{x\in X}{\arg \min}f_{LF}(x)$
        \State $\varepsilon^{eval}\gets f_{HL}(x^{eval})$
      \If{$\varepsilon^{eval} < \min \varepsilon^{itr}$}
        \State $c_{exploit}\gets c; c_{explore}\gets 0$
        \State $counter \gets 0$
        
      \Else
        \State $counter \gets counter + 1$
      \EndIf

      \If{$counter = 2$}
         \State $c_{exploit}\gets 0; c_{explore}\gets c$
      \EndIf
      
      \State Add $x^{eval}$ to $B^{itr}$ and $\varepsilon^{eval}$ to $\varepsilon^{itr}$
      
\EndFor

\State \textbf{return} $B^{itr},\varepsilon^{itr}$

\end{algorithmic}
\end{algorithm}

\section{Results}
\label{sec:results}

\subsection{ML model performance}

\begin{figure}[H]
    \centering
    \includegraphics[width=1.0
    \linewidth]{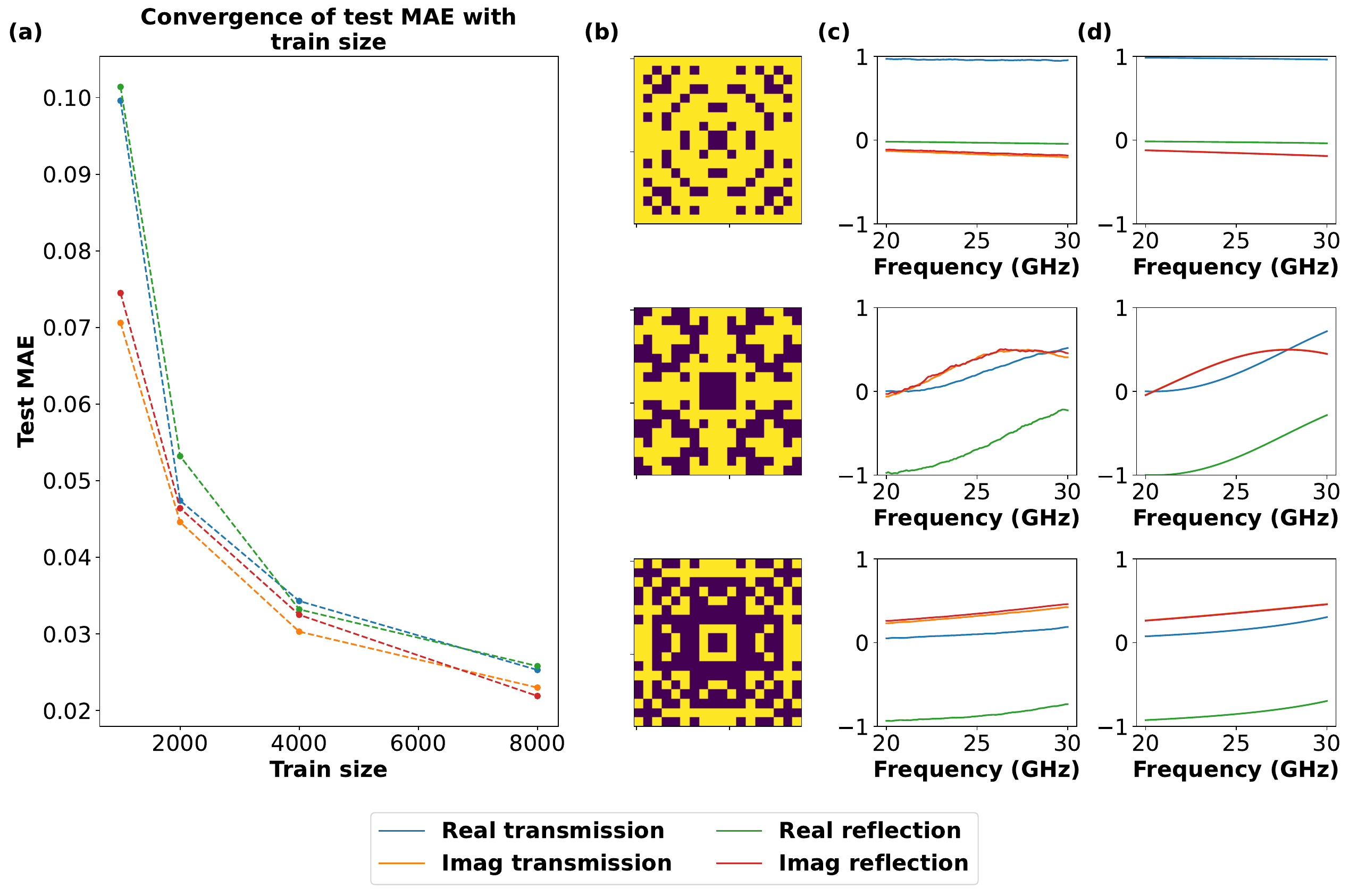}
    \caption{(a) Plot of Test MAE for the real and imaginary components of transmission and reflection from a surrogate model trained with increased training data. (b - d) Sample plots of real and imaginary transmission and reflection in the desired 20 - 30 GHz range for the metasurface designs in (b). (c) and (d) are ouputs from the surrogate model and the high-fidelity numerical solver respectively.}
    \label{fig:ml_vs_hfss}
\end{figure}

A deep learning model to predict responses of the frequency-selective surface in the 20 - 30 GHz range was previously trained. Test MAEs were calculated for models trained with training set sizes of 1000 to 8000 designs. Figure \ref{fig:ml_vs_hfss}(a) demonstrates a consistent improvement in model performance, as measured by test MAE, as the training dataset size increases. The surrogate model is able to provide average test MAE between 0.02 and 0.03 with 8000 training data points. Illustrative plots of predictions obtained by the trained neural network surrogate model and the ground truth obtained from high-fidelity numerical simulations are provided in Figure \ref{fig:ml_vs_hfss} for a few random metasurface designs in the test set. 

Even with an assumed 8-fold symmetry in meta-atom design of $18 \times 18$ discrete pixels, there are still $2^{45} [O(10^{13})]$ possible designs. Unsurprisingly, the performance does not yet asymptote at this training dataset size, which further reinforces that comprehensively sampling the design space \textit{a priori} is intractable and that there is a critical need for robust optimization in data-sparse regions. This highlights a fundamental challenge in AI-driven design: despite a significant computational investment (approximately two months) to generate the training data, the surrogate model remains far from converged. While more compute effort can be devoted to further dataset generation, this may not be optimal nor feasible for a design optimization problem as it is impractical for any \textit{a priori} collected training dataset to approach the full complexity of $2^{45}$ designs, especially in view of prior observations of a log-linear scaling in model performance with dataset size \cite{covert2024scaling}. This particular characteristic makes it critical to incorporate predictive uncertainty into the design workflow as the surrogate model is anticipated to have poorer predictive accuracy for truly novel designs encountered during the design optimization workflow due to inadequacies in the training data for extremely high-dimensional design problems. 

\subsection{Uncertainty metrics for neural network prediction}




Utilizing the test set (N = 1000), both the PHY-UNC and ENSB-UNC metrics were calculated for each design, and these predictions were compared to the prediction MAE from a base model. A Spearman's correlation coefficient was then calculated between each of the paired prediction MAE and the respective uncertainty measures in order to quantify how well each uncertainty metric correlates with the actual prediction error (MAE). Results for the PHY-UNC and ENSB-UNC are provided in Table \ref{tab: corr_physics}. For the ensemble approach, the models used share the same hyper-parameters, but are trained under different randomized initializations. 

\begin{table}[htbp]
\caption{Correlation between MAE and uncertainty estimate metric for different prediction models}
\begin{center}
\begin{tabular}{|c|c|c|}
\hline
\textbf{Component} & \textbf{ENSB-UNC} & \textbf{PHY-UNC} \\
\hline
$Re(R_{xx})$ & $0.70$ & $0.75$\\
\hline
$Im(R_{xx})$ & $0.54$ & $0.53$\\
\hline
$Re(T_{xx})$ & $0.75$ & $0.74$\\
\hline
$Im(T_{xx})$ & $0.57$ & $0.57$\\
\hline
\end{tabular}
\label{tab: corr_physics}
\end{center}
\end{table}

Both PHY-UNC and ENSB-UNC are strongly correlated with prediction MAE, with Spearman correlation coefficients ranging from 0.53 to 0.75. Critically, the physics-informed metric (PHY-UNC) achieves a correlation comparable to the conventional ensemble-based metric (ENSB-UNC). While ENSB-UNC can potentially improve further with a larger ensemble, this would entail a trade-off in terms of the increased computational cost of training more models and additional computational cost during evaluation in the optimization loop.

\subsection{Design optimization performance}
In the following Sections, results from different versions of the optimization process are presented. 100 independent runs are repeated for each scenario, and all optimizations are run for $N_{itr}=30$ iterations. For the rest of this analysis, we consider an optimized design as one where the final design achieves a mean absolute error of $<0.1$ as determined by the high-fidelity numerical solver (i.e., HF-DES-MAE $<0.1$). Hence, we also calculate and report the overall cumulative distribution function of the optimization algorithms' HF-DES-MAE across all 100 runs, and the fraction of these 100 runs that have achieved the target of HF-DES-MAE $<0.1$. Based on the cumulative distribution functions, we also do a Komolgorov-Smirnov test (KS-test) to determine if optimization results from the pair-wise scenarios are statistically different.


\subsubsection{Design performance with baseline BPSO} \label{sec:baseline_BPSO}
\begin{figure}[!t]
	\begin{center}
		{\includegraphics[width=1.0\columnwidth]{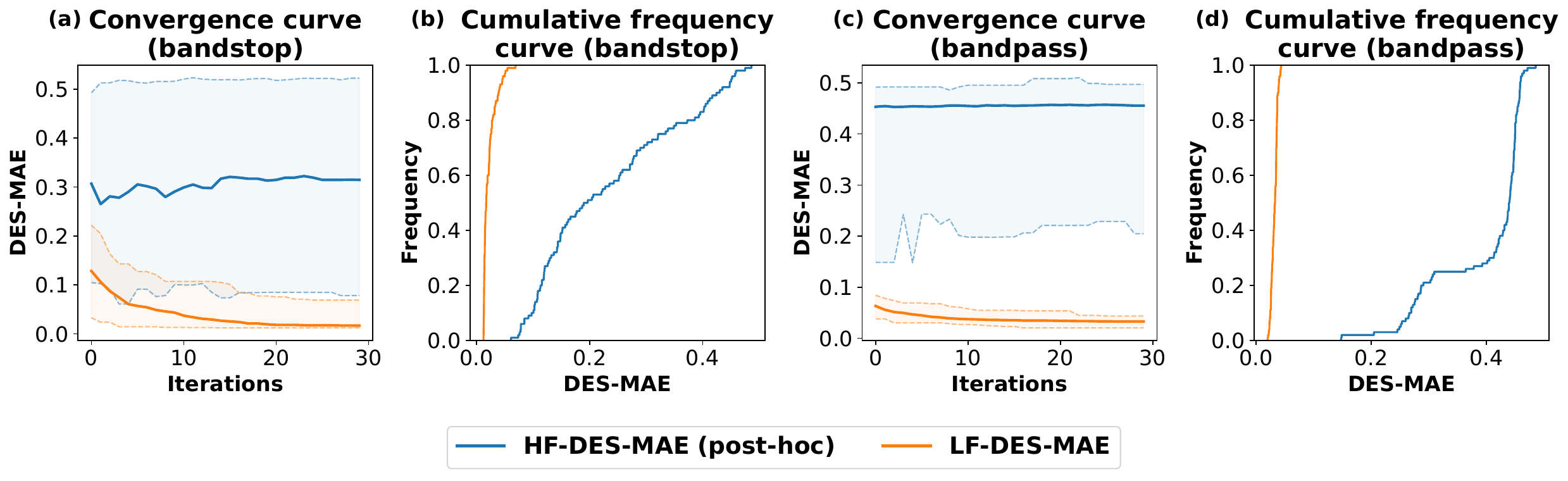}}
		\caption{Results of design optimization for the (a), (c) band-stop and (b), (d) band-pass profiles using a baseline BPSO algorithm where particle and global bests are evaluated based on LF-DES-MAE. (a), (c) Curves of the global best particle's HF-DES-MAE and LF-DES-MAE where LF-DES-MAE informs us about the ability of the BPSO to converge, whereas HF-DES-MAE indicates the performance of the putative best design during optimization. (b), (d) Cumulative frequency curves of the lowest HF-DES-MAE and LF-DES-MAE obtained at the end of each optimization run across 100 different runs.}
        \label{Fig:bpso_convergence}
	\end{center}
\end{figure}

Briefly, both particle and global best are evaluated by the surrogate model at each iteration (LF-DES-MAE) in the baseline BPSO implementation. The global bests are recorded at each iteration, and subsequently evaluated with the high-fidelity solver post-hoc (at the end of the design optimization). The convergence curves of the global bests' HF-DES-MAE and LF-DES-MAE for both the band-stop and band-pass profile are shown in Figure \ref{Fig:bpso_convergence}(a) and (c) while the corresponding cumulative frequency curve of the lowest LF-DES-MAE and HF-DES-MAE across different runs are presented in \ref{Fig:bpso_convergence}(b) and (d).

In both cases (band-stop and band-pass), the  LF-DES-MAE rapidly decreased across iterations and most converged to a low value of less than 0.05. While the optimization effectively identifies designs with low LF-DES-MAE, subsequent high-fidelity evaluation reveals that this apparent success is misleading. The corresponding HF-DES-MAE is much higher when evaluated post-hoc. As the training set has few training examples with $\text{HF-DES-MAE}$ $<0.05$ for both desired profiles, the surrogate model's prediction can be inaccurate, especially for potential novel and out-of-distribution designs encountered during the design optimization process. 

A key finding is that there is a significantly larger discrepancy between the LF-DES-MAE and corresponding HF-DES-MAE (evaluated post-hoc) for the band-pass profile than the band-stop profile. Figure~\ref{fig:profile_hist} clearly indicated the presence of more band-pass-type profiles in the design space. This suggests the surrogate model may be systematically biased towards predicting band-pass-like responses, even for designs outside its training distribution, leading the optimizer into false minima. 

\subsubsection{Poor design performance originates from poor surrogate predictions} \label{sec:single_metric_BPSO}

The results in Section \ref{sec:baseline_BPSO} suggest that poor surrogate model predictive performance is a primary cause of suboptimal design convergence. To test this hypothesis directly, we performed an analysis where the influence of the potentially erroneous global best is removed from the optimization. Unlike \ref{sec:baseline_BPSO} which keeps track of the particle with lowest LF-DES-MAE thus far, the particle with the lowest LF-DES-MAE is recorded at each iteration so that its transmission profile can be re-evaluated post-hoc using a high-fidelity solver (similar to Fig~\ref{Fig:bpso_convergence}). 

\begin{figure}[!t]
	\begin{center}
		{\includegraphics[width=1.0\columnwidth]{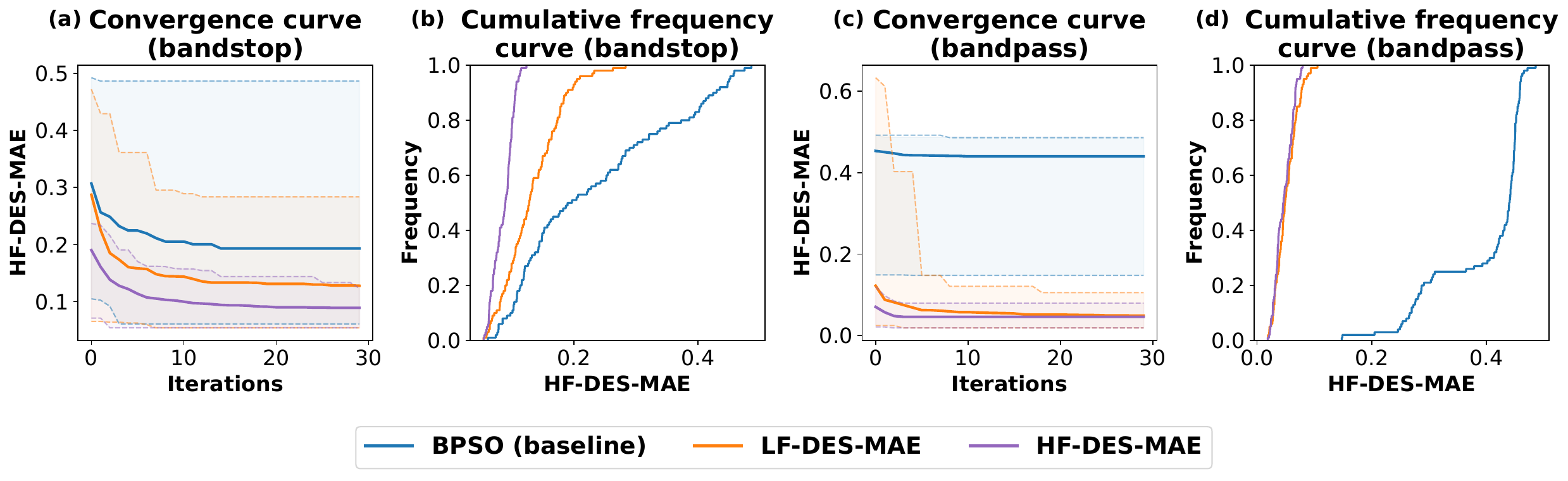}}
		\caption{Results of using baseline BPSO and using only particle bests as determined by HF-DES-MAE or LF-DES-MAE (a),(c) Convergence curve of the cumulative minimum (post-hoc evaluated) HF-DES-MAE over iterations (b), (d) Cumulative frequency curve of the lowest HF-DES-MAE obtained at the end of each optimization across 100 different runs}
        \label{Fig:single_pbest_curve_hf_lf}
	\end{center}
\end{figure}

The convergence curve of the cumulative minimum of the HF-DES-MAE is plotted in Figure \ref{Fig:single_pbest_curve_hf_lf}(a) and (c) and also indicate convergence within the 30 iterations of the optimization process. The cumulative frequency curves of the lowest HF-DES-MAE across different runs are presented in Figure \ref{Fig:single_pbest_curve_hf_lf}(b) and (d). The corresponding curves for the baseline BPSO (differentiated by its inclusion of guidance by a global best in the update step) are also included for comparison. 

Critically, the exclusion of guidance by a potentially erroneous global best (as indicated by `LF-DES-MAE' in Figure \ref{Fig:single_pbest_curve_hf_lf}) also significantly improves the robustness of the optimization process, as is consistent with the hypothesis. In addition, we repeat the experiment, but employ the high-fidelity solver to directly evaluate each particle at every iteration (`HF-DES-MAE' in  Figure \ref{Fig:single_pbest_curve_hf_lf}). Being the ideal scenario where there is no error in the assessment of the best design during each iteration in the optimization process, this establishes a performance upper bound as the most reliable example of convergence towards the target design. Across both the band-stop and band-pass scenarios, this approach achieves the highest fraction of optimized designs (defined as HF-DES-MAE $< 0.1$). 

 Interestingly, there is a difference in the performance for the band-stop and band-pass target. In the optimization towards a band-stop profile, the sole use of a low-fidelity surrogate model had significantly worse optimization performance than the ideal case where the high-fidelity solver is used for all particles. In contrast, both strategies displayed similar performance for the band-pass profile. This may be attributed to band-pass profiles being more prevalent in the design space, hence, the optimization process is still able to identify designs within the design space encapsulated in its original training dataset, and thereby identify performant band-pass designs. 



\subsubsection{Design performance with uncertainty-aware BPSO}

\begin{figure}[!t]
	\begin{center}
		{\includegraphics[width=1.0\columnwidth]{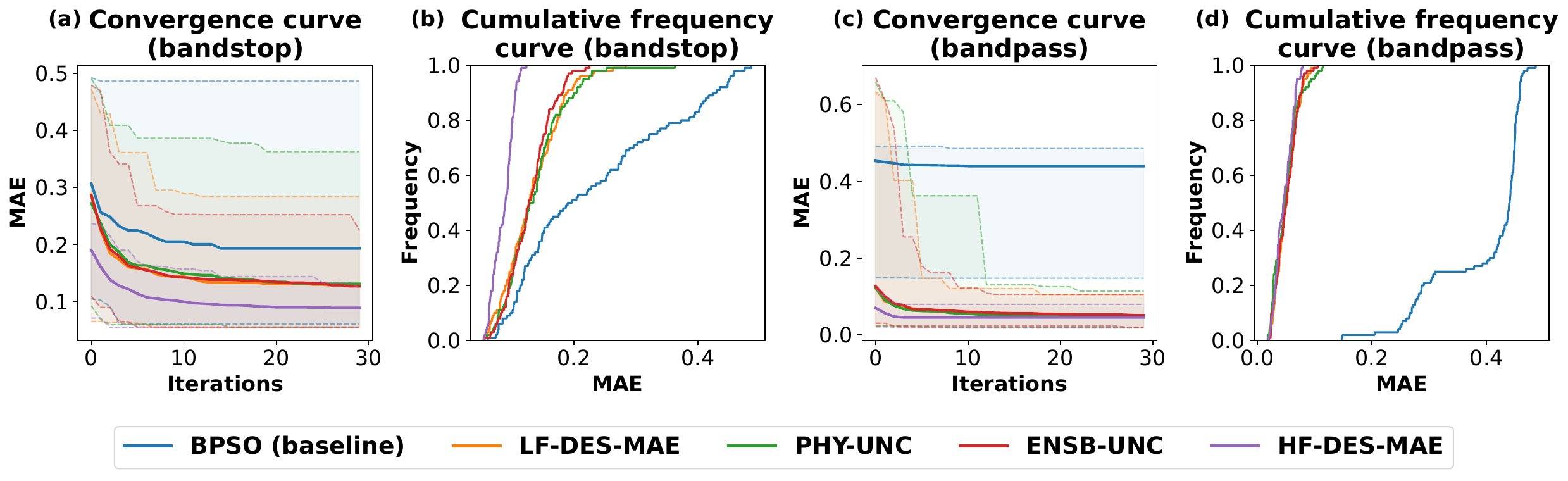}}
		\caption{Results of using particle best updated with a single metric (PHY-UNC, ENSB-UNC, LF-DES-MAE, HF-DES-MAE) (a),(c) Convergence curve of the cumulative minimum HF-MAE over iterations (b), (d) Cumulative frequency curve of the lowest HF-DES-MAE across different runs}
        \label{Fig:single_pbest_curve}
	\end{center}
\end{figure}

As a further comparison, we repeat the experiment, but utilize uncertainty metrics (PHY-UNC and ENSB-UNC) as a means of identifying potential designs. Figure \ref{Fig:single_pbest_curve} provides a visualization of the (a, c) convergence behavior and (b, d) final outcome distribution across different optimization runs for both the band-stop and band-pass profiles.

As tabulated in Table \ref{tab:single_metric_robust}, the use of the high-fidelity numerical simulation for evaluation during optimization yields the best results, with $100\%$ and $79\%$ of the optimization runs for the band-pass and band-stop target achieving HF-DES-MAE $<0.1$ respectively. Interestingly, the performance when incorporating the uncertainty metrics was similar to the use of LF-DES-MAE, with all 3 scenarios having approximately 99\% and 25\% of the band-pass and band-stop optimization runs yielding designs with HF-DES-MAE $<0.1$. For this design optimization problem, the outcomes are similar when a low-fidelity surrogate is used (thereby driving the optimization to data-poor regions where prediction uncertainty is high, although performance is predicted to be good), and when the particles are directly being driven to high uncertainty regions in the design space. This indicates that the exploration of high uncertainty regions can yield similar beneficial effects to performance guidance by the surrogate model for the design optimization process.

\begin{table}[]
    \centering
    \begin{tabular}{ c c c } \hline 
     & Band-stop& Band-pass\\ \hline 
     BPSO& $10$&$0$\\ \hline
     HF-DES-MAE& $79$&$100$\\
     LF-DES-MAE& $28$&$99$\\ 
     ENSB-UNC& $24$& $99$\\  
     PHY-UNC & $23$& $96$\\  \hline
    \end{tabular}
    \caption{Percentage of runs with best HF-DES-MAE below 0.1}
    \label{tab:single_metric_robust}
\end{table}

Results of the Kolmogorov-Smirnov (KS) test are presented in Table \ref{tab: single_metric} for the different pair-wise permutations for both band-pass and band-stop target profiles. From Table \ref{tab: single_metric}, the baseline BPSO is consistently worse than all the other algorithms (statistically significant with $p<0.05$), while the HF-DES-MAE is consistently better than all the other algorithms for the band-stop problem (statistically significant with $p<0.05$), and similar in performance for the band-pass problem to scenarios where PHY-UNC, ENSB-UNC or LF-DES-MAE are used as the metric to update the particle bests. 

\begin{table}[!t]
    \centering
    \begin{tabular}{c c c c } \hline 
         Model 1& Model 2& Band-stop& Band-pass\\ \hline  
         BPSO& HF-DES-MAE& + & +\\  
         BPSO& LF-DES-MAE& + & +\\  
         BPSO& PHY-UNC& + & +\\  
         BPSO& ENSB-UNC& + & +\\ \hline
         HF-DES-MAE& LF-DES-MAE& + & -\\  
         HF-DES-MAE& PHY-UNC& + & -\\ 
         HF-DES-MAE& ENSB-UNC& + & -\\ \hline  
         LF-DES-MAE& PHY-UNC& - & -\\ 
         LF-DES-MAE& ENSB-UNC& - & -\\ \hline
         PHY-UNC& ENSB-UNC& - & -\\\hline\end{tabular}
        \caption{Measure of how likely it is that the cumulative distribution curves of their optimization outcomes come from the same underlying probability distribution as determined by the p-values from a Kolmogorov-Smirnov test. + represents p $<0.05$ (i.e. likely to be statistically dissimilar) while - represents p $>0.05$.}
        \label{tab: single_metric}
\end{table}

Importantly, the difference in optimization performance between PHY-UNC and ENSB-UNC is also not statistically significant, indicating that the PHY-UNC metric is as effective as ENSB-UNC when incorporated into the equivalent design optimization workflow. 



\subsubsection{Design performance with Multi-fidelity Uncertainty-based BPSO}

\begin{figure}[!t]
	\begin{center}
		\includegraphics[width=1.0\columnwidth]{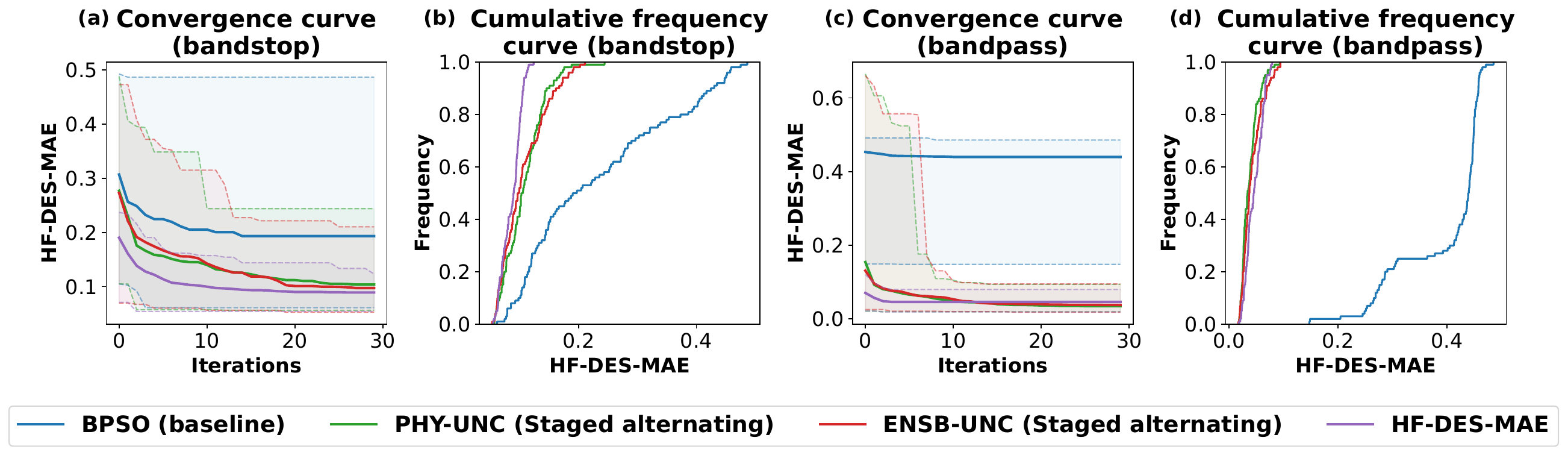}
		\caption{Results of alternating between exploitation and exploration (a),(c) Convergence curve of the cumulative minimum MAE over iterations (b), (d) Cumulative frequency curve of the lowest MAE across different runs}
           \label{fig:alternation}
	\end{center}
\end{figure}

While the individual uncertainty metrics are shown to improve exploration, single-fidelity optimisation does not fully leverage information from the costly high-fidelity evaluations. To bridge this gap, we developed a multi-fidelity framework that actively incorporates high-fidelity results into the search process to balance exploration and exploitation. Figure \ref{fig:alternation} provides a visualization of the (a, c) convergence behavior and (b, d) final outcome distribution across different optimization runs for both the band-stop and band-pass profiles.

The multi-fidelity uncertainty-aware BPSO demonstrated a substantial improvement in optimization performance over all previous methods. As shown in Figure~\ref{fig:alternation} and Table \ref{tab:sig_alternation_robustness}, this multi-fidelity approach dramatically increased the success rate, with $\approx50\%$ of runs for the challenging band-stop profile and 100\% of runs for the band-pass profile finding designs with a discrepancy of less than $0.1$ from the target profile (HF-DES-MAE). This method obtained results closer to the baseline situation where a high-fidelity solver was directly used for every particle evaluation during the design workflow. 

\begin{table}[]
    \centering
    \begin{tabular}{ c c c } \hline 
     & Band-stop& Band-pass\\ \hline  
     PHY-UNC (staged)& $45$& $100$\\  
     ENSB-UNC (staged)& $52$& $100$\\  
     \hline
    \end{tabular}
    \caption{Percentage of runs with best HF-DES-MAE below 0.1}
    \label{tab:sig_alternation_robustness}
\end{table}

Based on a KS-test, the cumulative distribution curves of the optimization outcomes from the multi-fidelity uncertainty-aware optimization process using PHY-UNC or ENSB-UNC were also shown to not be statistically significant for both the band-pass and band-stop target profile ($p > 0.05$). This further highlights that the PHY-UNC is able to function as well as the more conventional ENSB-UNC.


\subsubsection{Comparison of optimization runtimes}

A key benefit of this method is the reduction in time taken for the optimization. The most time-consuming part of this process is the high-fidelity solver evaluation. Table \ref{tab:runtime} details the time taken for each of the algorithms proposed in this work to complete one run which produces one design. For consistency, 10 iterations of the constant phase and 20 iterations of the alternating phase are performed for the multi-fidelity uncertainty-aware BPSO (for both PHY-UNC and ENSB-UNC), while 30 iterations are performed for all other BPSO variants.

Using a multi-fidelity approach for optimization reduces the time taken through a judicious selection of particles for high-fidelity evaluation. Performing 100 runs using the multi-fidelity PHY-UNC-based PSO algorithm took around 4 hours whereas 100 optimization runs with evaluation of all particles by the high-fidelity solver (HF-DES-MAE) took multiple days. The significant reduction in runtime was achieved by performing the high-fidelity evaluation on only one particle per iteration instead of all particles. Critically, an enhanced robustness can also be achieved despite this dramatic increase in computational efficiency. As detailed in Table \ref{tab:runtime}, the multi-fidelity PHY-UNC algorithm reduces the runtime per optimization run by 10x compared to the full high-fidelity algorithm (158.0 s vs. 1419.1 s), while maintaining a statistically indistinguishable optimization outcome for the band-pass target ($100\%$ of the optimization runs achieved HF-DES-MAE $<0.1$). 

\begin{table}
    \centering
    \begin{tabular}{cc}
    \hline
         & Runtime (s)\\
         \hline
         BPSO& $35.0\pm 1.6$\\ \hline
         HF-DES-MAE& $1419.1\pm318.5$\\
         LF-DES-MAE& $35.8\pm 0.6$\\
         PHY-UNC& $107.6\pm 2.1$\\
         ENSB-UNC& $389.8\pm2.0$\\ \hline
         PHY-UNC (staged)& $158.0\pm 5.4$\\
         ENSB-UNC (staged)& $443.5\pm 2.4$\\
    \hline
    \end{tabular}
    \caption{Runtime of different algorithms for a single run with 30 iterations}
    \label{tab:runtime}
\end{table}

\section{Discussion}
\label{sec:discuss}

Overall, our experiments revealed that integrating uncertainty measures within a surrogate-assisted design workflow offers substantial advantages. The results clearly demonstrate that incorporation of uncertainty measures significantly improves robustness, even while retaining benefits in accelerated design convergence over conventional use of high-fidelity numerical solvers. In our experiments, performant designs could be obtained in 10-fold less time compared to the use of high-fidelity solvers, and with \textbf{$4.5$}-fold increase in reliability compared to the conventional use of surrogate models. 

The enhanced reliability stems from the algorithm's ability to avoid excessively optimising towards regions in the design space where the surrogate model is prone to inaccurate predictions. By re-evaluating designs using a high-fidelity solver, the algorithm mitigates the risk of converging to suboptimal solutions based on erroneous surrogate predictions. This is particularly crucial for novel designs, where the surrogate model may not have sufficient training data to accurately predict performance.


A key innovation in this work is the demonstrated potential for estimating predictive uncertainty via leveraging fundamental physics principles, specifically the continuity in the electromagnetic field in this case. Physics Uncertainty (PHY-UNC) provides a robust indicator of potential errors in the surrogate model's predictions, effectively guiding the optimization algorithm to explore regions of the unexplored design space where novel designs may exist, and to selectively re-evaluate the designs in those regions that are most likely to be performant as assessed by the lowest LF-DES-MAE. In contrast to other data-driven uncertainty measures (e.g. ensembles), PHY-UNC incorporates domain expertise, and can be a computationally cheaper alternative while being potentially as effective. 

While the PHY-UNC metric in this work is derived from a fundamental, yet simple physical law involving transmission and reflection of the electromagnetic field, its core concept of using violations of known physical principles as a proxy for uncertainty is broadly generalizable. For example, PHY-UNC could potentially be useful in other multi-physics systems by incorporating partial domain-specific empirical laws, or other conservation equations derived from governing PDEs. A simple example is to check conservation of mass in surrogate predictions of a flow field in fluid dynamics (i.e. the continuity equation). Nonetheless, its limits in applicability to more complex systems and relations should be more thoroughly investigated in future work. 

In fact, domain experts could play a critical role in this process by prescribing additional soft-constraint terms or interpretable physics rules that capture known behaviors (e.g., symmetry, monotonicity, conservation properties) even if they do not stem from closed-form equations. This modularity suggests that PHY-UNC could serve as a generalizable and customizable uncertainty framework which is adaptable to new domains and physical systems beyond electromagnetics as demonstrated in this work. 

\section{Conclusion}
\label{sec:conclusion}

In this work, we demonstrated an uncertainty-aware framework for the inverse design of frequency-selective surfaces, leveraging either deep ensemble or physics-based principles for uncertainty quantification. By integrating uncertainty into the surrogate-assisted optimization loop, we mitigate the risks of surrogate model extrapolation errors and significantly improve the reliability of the design workflow. 

Our results show that the proposed method not only accelerates the optimization process by reducing reliance on high-fidelity solvers but also enhances confidence in the final design outcomes. The ability to estimate uncertainty enables a more informed exploration of the design space, making this approach particularly valuable for engineering applications where high-dimensional, data-limited design problems are prevalent. 

Nonetheless, it is important to acknowledge the limitations of the current study and potential directions for extension. While the proposed physical principle used here demonstrated remarkable effectiveness, it may be interesting to develop more sophisticated uncertainty measures that incorporate additional physics principles or leverage other types of domain knowledge and evaluate their relative performance. In addition, other design workflows are being developed that utilize active learning strategies to adaptively expand the training dataset in regions where the surrogate model is most uncertain, targeting the construction of larger, more effective datasets that can potentially mitigate the issue of poor performance from the surrogate model. Nonetheless, we believe that the ability to estimate uncertainty via domain knowledge can still be of utility in automated, high-throughput workflows to ward against unanticipated poor model performance.



\clearpage

\appendix

\section{Algorithm flow charts} \label{sec:flowcharts}
Illustrative flow charts of the algorithm flows are presented here. 

\begin{figure}[bthp]
	\begin{center}
		{\includegraphics[width=1.0\columnwidth]{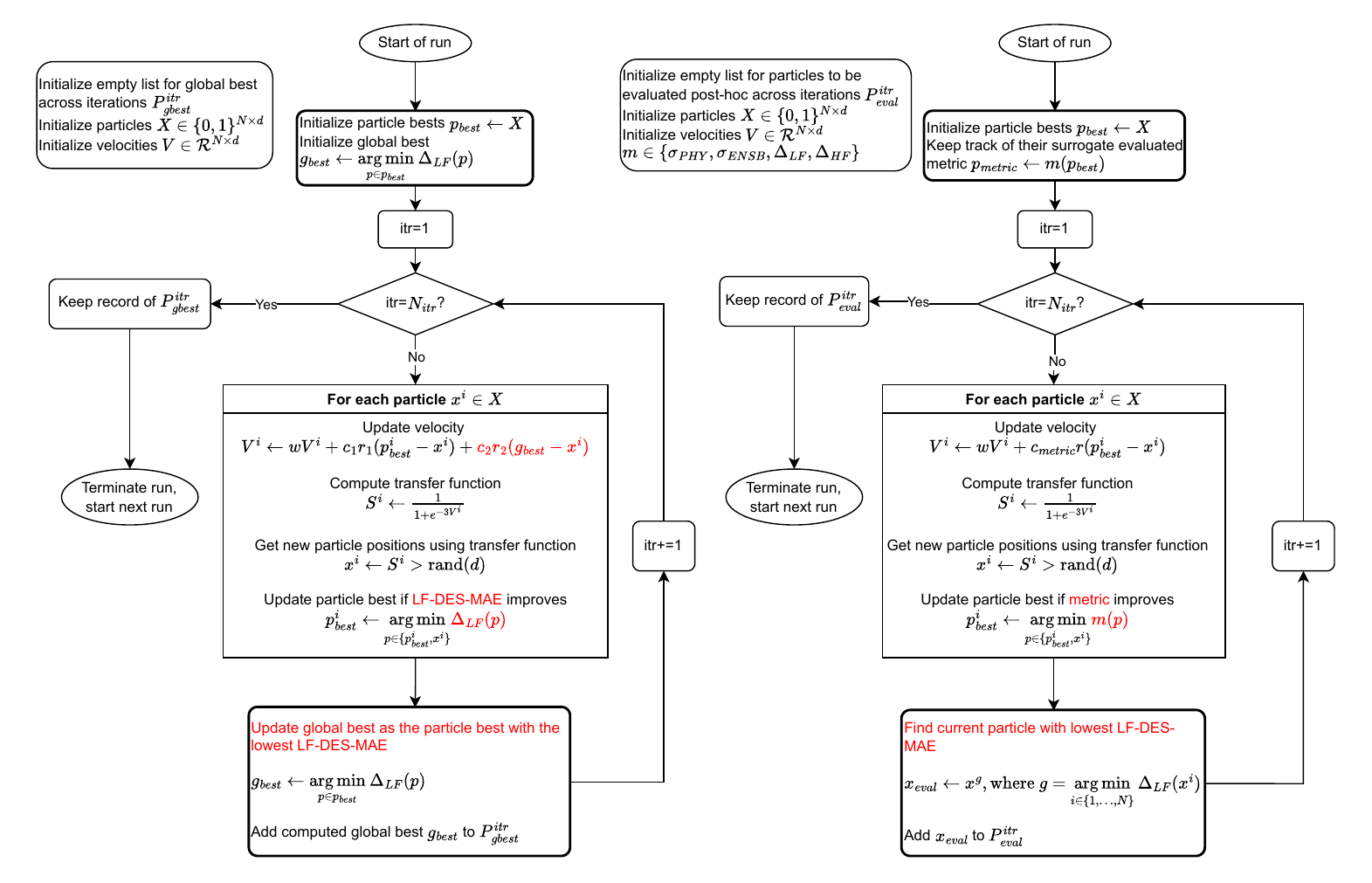}}
		\caption{Flow chart showing algorithm for the baseline BPSO algorithm and a particle swarm optimization algorithm without the global best metric}\label{Fig:algorithm_flowchart_1v2}
	\end{center}
\end{figure}

Figure \ref{Fig:algorithm_flowchart_1v2} illustrates how the key difference between the baseline BPSO algorithm and the algorithms detailed in Section \ref{sec:single_metric_BPSO} is the removal of guidance by the global best during the optimization process.

\begin{figure}[bthp]
	\begin{center}
		{\includegraphics[width=1.0\columnwidth]{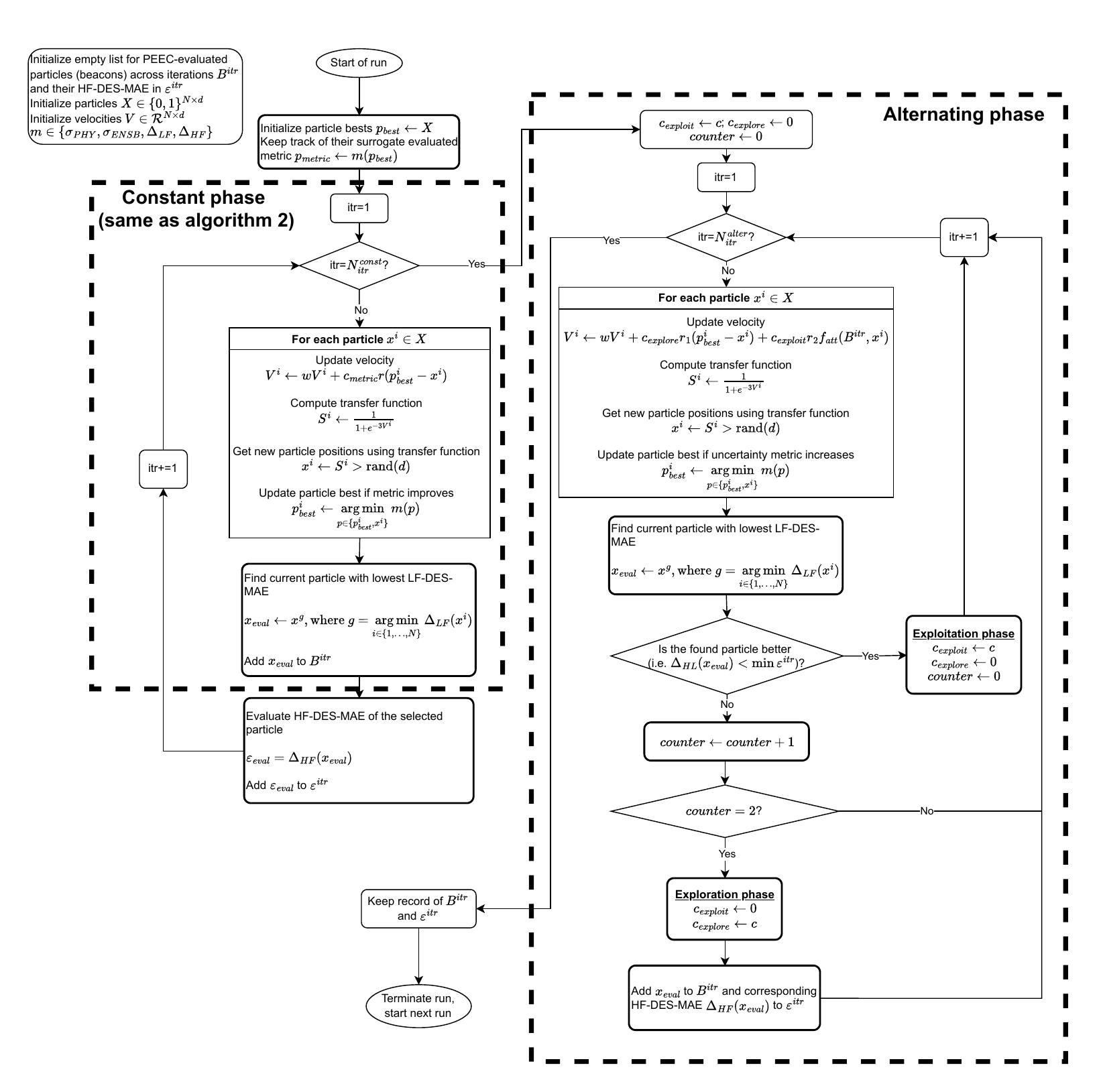}}
		\caption{Flow chart showing the two-stage multi-fidelity optimization algorithm with the constant phase and the alternating phase}\label{Fig:algorithm_flowchart_3}
	\end{center}
\end{figure}

Figure \ref{Fig:algorithm_flowchart_3} illustrates how the multi-fidelity BPSO algorithm consists of a two-stage process, with 10 iterations in the `constant-attraction' stage, and a final 20 iterations in the `alternating-strategy' stage which seeks to balance exploration and exploitation during the optimization. The algorithm is detailed in Section \ref{sec:multifidelity_bpso}.

\clearpage

\section{Staged vs Purely Alternating}
\begin{figure}[H]
    \centering
    \includegraphics[width=1.0\linewidth]{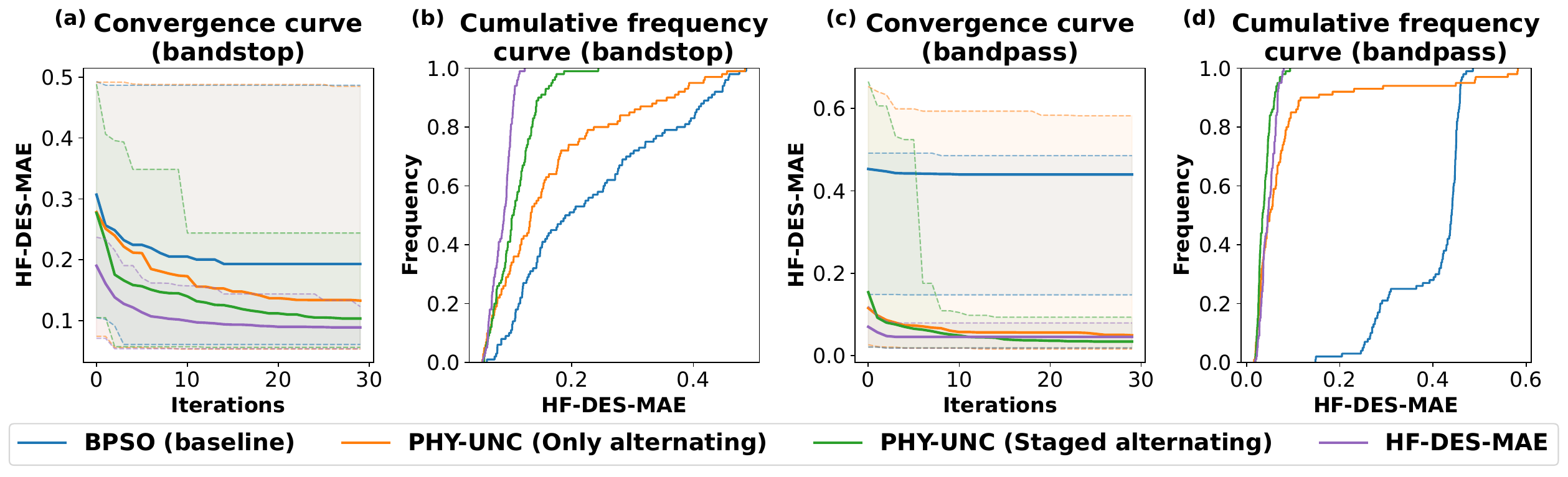}
    \caption{Comparison of performance between having all 30 iterations use the `alternating-strategy' (PHY-UNC (Only alternating)) and having 10 iterations of `constant-attraction' prior to 20 iterations of `alternating-strategy' (PHY-UNC (Staged alternating))}
    \label{fig:only_vs_staged}
\end{figure}

Figure \ref{fig:only_vs_staged} shows the reason for having 10 iterations of `constant-attraction' prior to 20 iterations of `alternating-strategy' instead of the sole use of the `alternating-strategy'. From the convergence curves, it can be observed that the `alternating-strategy' led to the intended result of higher variance in the optimization process. However, this also meant that more runs are unsuccessful at obtaining performant designs (some designs still have HF-DES-MAE above 0.5). From Figure \ref{Fig:single_pbest_curve}, it can be observed that for a large number of runs, HF-DES-MAE stopped reducing after around 10 iterations of using PHY-UNC as the single metric for optimization. We demonstrate that applying the `alternating-strategy' after this initial drop (`cold-start') is beneficial.

\clearpage

\bibliographystyle{elsarticle-num} 
\bibliography{cas-refs}

@article{ma2021deep,
  title={Deep learning for the design of photonic structures},
  author={Ma, Wei and Liu, Zhaocheng and Kudyshev, Zhaxylyk A and Boltasseva, Alexandra and Cai, Wenshan and Liu, Yongmin},
  journal={Nature Photonics},
  volume={15},
  number={2},
  pages={77--90},
  year={2021},
  publisher={Nature Publishing Group UK London}
}

@article{lakshminarayanan2017simple,
  title={Simple and scalable predictive uncertainty estimation using deep ensembles},
  author={Lakshminarayanan, Balaji and Pritzel, Alexander and Blundell, Charles},
  journal={Advances in neural information processing systems},
  volume={30},
  year={2017}
}

@article{tran2019bayesian,
  title={Bayesian layers: A module for neural network uncertainty},
  author={Tran, Dustin and Dusenberry, Mike and Van Der Wilk, Mark and Hafner, Danijar},
  journal={Advances in neural information processing systems},
  volume={32},
  year={2019}
}

@article{wenzel2020hyperparameter,
  title={Hyperparameter ensembles for robustness and uncertainty quantification},
  author={Wenzel, Florian and Snoek, Jasper and Tran, Dustin and Jenatton, Rodolphe},
  journal={Advances in Neural Information Processing Systems},
  volume={33},
  pages={6514--6527},
  year={2020}
}

@article{calik2021accurate,
  title={Accurate modeling of frequency selective surfaces using fully-connected regression model with automated architecture determination and parameter selection based on Bayesian optimization},
  author={Calik, Nurullah and Belen, Mehmet Ali and Mahouti, Peyman and Koziel, Slawomir},
  journal={IEEE Access},
  volume={9},
  pages={38396--38410},
  year={2021},
  publisher={IEEE}
}

@article{fontoura2021synthesis,
  title={Synthesis of multiband frequency selective surfaces using machine learning with the decision tree algorithm},
  author={Fontoura, Leidiane CMM and Lins, Hertz Wilton De Castro and Bertuleza, Arthur S and D’assun{\c{c}}{\~a}o, Adaildo Gomes and Neto, Alfredo Gomes},
  journal={IEEE Access},
  volume={9},
  pages={85785--85794},
  year={2021},
  publisher={IEEE}
}

@article{genovesi2006particle,
  title={Particle swarm optimization for the design of frequency selective surfaces},
  author={Genovesi, Simone and Mittra, Raj and Monorchio, Agostino and Manara, Giuliano},
  journal={IEEE Antennas and Wireless Propagation Letters},
  volume={5},
  pages={277--279},
  year={2006},
  publisher={IEEE}
}

@inproceedings{yang2022physics,
  title={Physics Compliance as a Metric for Neural Network Uncertainty},
  author={Yang, Zhong Liang Ou and Jiang, Yang and Wong, Jian Cheng and Chiu, Pao--Hsiung and Zhao, Weijiang and Dao, My Ha and Ooi, Chin Chun},
  booktitle={2022 IEEE Symposium Series on Computational Intelligence (SSCI)},
  pages={01--07},
  year={2022},
  organization={IEEE}
}

@inproceedings{kennedy1997discrete,
  title={A discrete binary version of the particle swarm algorithm},
  author={Kennedy, James and Eberhart, Russell C},
  booktitle={1997 IEEE International conference on systems, man, and cybernetics. Computational cybernetics and simulation},
  volume={5},
  pages={4104--4108},
  year={1997},
  organization={ieee}
}

@article{mirjalili2013s,
  title={S-shaped versus V-shaped transfer functions for binary particle swarm optimization},
  author={Mirjalili, Seyedali and Lewis, Andrew},
  journal={Swarm and Evolutionary Computation},
  volume={9},
  pages={1--14},
  year={2013},
  publisher={Elsevier}
}

@article{jiang2022full,
  title={A full-wave generalized PEEC model for periodic metallic structure with arbitrary shape},
  author={Jiang, Yang and Zhao, Wei-Jiang and Gao, Richard Xian-Ke and Liu, En-Xiao and Png, Ching Eng},
  journal={IEEE Transactions on Microwave Theory and Techniques},
  volume={70},
  number={9},
  pages={4110--4119},
  year={2022},
  publisher={IEEE}
}

@article{jiang2022compact,
  title={Compact quasi-static PEEC modeling of electromagnetic problems with finite-sized dielectrics},
  author={Jiang, Yang and Gao, Richard Xian-Ke},
  journal={IEEE Transactions on Microwave Theory and Techniques},
  volume={71},
  number={6},
  pages={2373--2383},
  year={2022},
  publisher={IEEE}
}

@article{jiang2023peec,
  title={PEEC Model Based on a Novel Quasi-Static Green's Function for Two-Dimensional Periodic Structures},
  author={Jiang, Yang and Dou, Yuhang and Gao, Richard Xian-Ke},
  journal={IEEE Journal on Multiscale and Multiphysics Computational Techniques},
  volume={8},
  pages={187--194},
  year={2023},
  publisher={IEEE}
}

@article{zagoruyko2016wide,
  title={Wide residual networks},
  author={Zagoruyko, Sergey and Komodakis, Nikos},
  journal={arXiv preprint arXiv:1605.07146},
  year={2016}
}

@article{karniadakis2021physics,
  title={Physics-informed machine learning},
  author={Karniadakis, George Em and Kevrekidis, Ioannis G and Lu, Lu and Perdikaris, Paris and Wang, Sifan and Yang, Liu},
  journal={Nature Reviews Physics},
  volume={3},
  number={6},
  pages={422--440},
  year={2021},
  publisher={Nature Publishing Group UK London}
}

@article{xie2023inverse,
  title={Inverse design of chiral functional films by a robotic AI-guided system},
  author={Xie, Yifan and Feng, Shuo and Deng, Linxiao and Cai, Aoran and Gan, Liyu and Jiang, Zifan and Yang, Peng and Ye, Guilin and Liu, Zaiqing and Wen, Li and others},
  journal={Nature Communications},
  volume={14},
  number={1},
  pages={6177},
  year={2023},
  publisher={Nature Publishing Group UK London}
}

@article{li2022empowering,
  title={Empowering metasurfaces with inverse design: principles and applications},
  author={Li, Zhaoyi and Pestourie, Rapha{\"e}l and Lin, Zin and Johnson, Steven G and Capasso, Federico},
  journal={Acs Photonics},
  volume={9},
  number={7},
  pages={2178--2192},
  year={2022},
  publisher={ACS Publications}
}

@article{yang2023inverse,
  title={Inverse design optimization framework via a two-step deep learning approach: application to a wind turbine airfoil},
  author={Yang, Sunwoong and Lee, Sanga and Yee, Kwanjung},
  journal={Engineering with Computers},
  volume={39},
  number={3},
  pages={2239--2255},
  year={2023},
  publisher={Springer}
}

@article{li2018metasurfaces,
  title={Metasurfaces and their applications},
  author={Li, Aobo and Singh, Shreya and Sievenpiper, Dan},
  journal={Nanophotonics},
  volume={7},
  number={6},
  pages={989--1011},
  year={2018},
  publisher={De Gruyter}
}

@article{naseri2021combined,
  title={A combined machine-learning/optimization-based approach for inverse design of nonuniform bianisotropic metasurfaces},
  author={Naseri, Parinaz and Pearson, Stewart and Wang, Zhengzheng and Hum, Sean V},
  journal={IEEE Transactions on Antennas and Propagation},
  volume={70},
  number={7},
  pages={5105--5119},
  year={2021},
  publisher={IEEE}
}

@article{chatterjee2019critical,
  title={A critical review of surrogate assisted robust design optimization},
  author={Chatterjee, Tanmoy and Chakraborty, Souvik and Chowdhury, Rajib},
  journal={Archives of Computational Methods in Engineering},
  volume={26},
  pages={245--274},
  year={2019},
  publisher={Springer}
}

@article{li2019surrogate,
  title={Surrogate model uncertainty quantification for reliability-based design optimization},
  author={Li, Mingyang and Wang, Zequn},
  journal={Reliability Engineering \& System Safety},
  volume={192},
  pages={106432},
  year={2019},
  publisher={Elsevier}
}

@article{xiong2019data,
  title={Data-driven design space exploration and exploitation for design for additive manufacturing},
  author={Xiong, Yi and Duong, Pham Luu Trung and Wang, Dong and Park, Sang-In and Ge, Qi and Raghavan, Nagarajan and Rosen, David W},
  journal={Journal of Mechanical Design},
  volume={141},
  number={10},
  pages={101101},
  year={2019},
  publisher={American Society of Mechanical Engineers}
}

@article{jung2021confidence,
  title={Confidence-based design optimization for a more conservative optimum under surrogate model uncertainty caused by Gaussian process},
  author={Jung, Yongsu and Kang, Kyeonghwan and Cho, Hyunkyoo and Lee, Ikjin},
  journal={Journal of Mechanical Design},
  volume={143},
  number={9},
  pages={091701},
  year={2021},
  publisher={American Society of Mechanical Engineers}
}

@book{jiang2020surrogate,
  title={Surrogate-model-based design and optimization},
  author={Jiang, Ping and Zhou, Qi and Shao, Xinyu and Jiang, Ping and Zhou, Qi and Shao, Xinyu},
  year={2020},
  publisher={Springer}
}

@article{zhao2022limitations,
  title={Limitations of machine learning models when predicting compounds with completely new chemistries: possible improvements applied to the discovery of new non-fullerene acceptors},
  author={Zhao, Zhi-Wen and Del Cueto, Marcos and Troisi, Alessandro},
  journal={Digital Discovery},
  volume={1},
  number={3},
  pages={266--276},
  year={2022},
  publisher={Royal Society of Chemistry}
}

@article{miller2019predictive,
  title={Predictive abilities of machine learning techniques may be limited by dataset characteristics: insights from the UNOS database},
  author={Miller, P Elliott and Pawar, Sumeet and Vaccaro, Benjamin and McCullough, Megan and Rao, Pooja and Ghosh, Rohit and Warier, Prashant and Desai, Nihar R and Ahmad, Tariq},
  journal={Journal of cardiac failure},
  volume={25},
  number={6},
  pages={479--483},
  year={2019},
  publisher={Elsevier}
}

@article{raabe2023accelerating,
  title={Accelerating the design of compositionally complex materials via physics-informed artificial intelligence},
  author={Raabe, Dierk and Mianroodi, Jaber Rezaei and Neugebauer, J{\"o}rg},
  journal={Nature Computational Science},
  volume={3},
  number={3},
  pages={198--209},
  year={2023},
  publisher={Nature Publishing Group US New York}
}

@article{hawkins2021attractors,
  title={Attractors in dynamical systems},
  author={Hawkins, Jane and Hawkins, Jane},
  journal={Ergodic dynamics: From basic theory to applications},
  pages={27--39},
  year={2021},
  publisher={Springer}
}

@article{covert2024scaling,
  title={Scaling laws for the value of individual data points in machine learning},
  author={Covert, Ian and Ji, Wenlong and Hashimoto, Tatsunori and Zou, James},
  journal={arXiv preprint arXiv:2405.20456},
  year={2024}
}

@article{katwe2024overview,
  title={An overview of intelligent meta-surfaces for 6G and beyond: opportunities, trends, and challenges},
  author={Katwe, Mayur V and Kaushik, Aryan and Mohjazi, Lina and Abualhayja'a, Mohammad and Dardari, Davide and Singh, Keshav and Imran, Muhammad Ali and Butt, M Majid and Dobre, Octavia A},
  journal={IEEE Communications Standards Magazine},
  volume={8},
  number={4},
  pages={62--69},
  year={2024},
  publisher={IEEE}
}

@article{chen2024metasurfaces,
  title={Metasurfaces Empowering 6G Communication and Sensing: Opportunities and Challenges},
  author={Chen, Wenxiong and Chen, Lili and Zhao, Yizhe and Ren, Ju and Shen, Xuemin Sherman},
  journal={IEEE Wireless Communications},
  year={2024},
  publisher={IEEE}
}

@article{wang2024designing,
  title={Designing broadband cross-polarization conversion metasurfaces using binary particle swarm optimization algorithm},
  author={Wang, Jiao and Gu, Wei-Qi and Zhao, Xin-Cheng and Jiang, Yan-Nan and Xu, Kai-Da},
  journal={Materials \& Design},
  volume={247},
  pages={113419},
  year={2024},
  publisher={Elsevier}
}

@article{lalbakhsh2016multiobjective,
  title={Multiobjective particle swarm optimization to design a time-delay equalizer metasurface for an electromagnetic band-gap resonator antenna},
  author={Lalbakhsh, Ali and Afzal, Muhammad U and Esselle, Karu P},
  journal={IEEE Antennas and Wireless Propagation Letters},
  volume={16},
  pages={912--915},
  year={2016},
  publisher={IEEE}
}





\end{document}